\documentclass[runningheads]{llncs}

\usepackage[T1]{fontenc}
\usepackage{graphicx}
\usepackage{url}
\usepackage{comment}
\usepackage{amsmath,amssymb,graphicx}

\newcommand{\method}{{\sc $\gamma$-Quant}}
\usepackage[dvipsnames]{xcolor}

\usepackage{booktabs}
\usepackage{algpseudocode}
\usepackage{algorithm}
\usepackage{multirow}
\usepackage{multicol}
\usepackage{mathtools}
\usepackage{ragged2e}
\usepackage{minted}
\usepackage[export]{adjustbox}
\usepackage{soul}

\usepackage[capitalize]{cleveref}
\crefname{section}{Sec.}{Secs.}
\Crefname{section}{Section}{Sections}
\Crefname{table}{Table}{Tables}
\crefname{table}{Tab.}{Tabs.}
\definecolor{cadmiumgreen}{rgb}{0.0, 0.42, 0.24}
\definecolor{custom}{cmyk}{0.1,0.48,0.49,0.2}
\definecolor{OliveGreen}{cmyk}{0.64,0,0.95,0.40}
\definecolor{new}{rgb}{0.81,0.05,0.9}
\definecolor{BrickRed}{rgb}{0.81,0.1,0.1}
\definecolor{RoyalBlue}{rgb}{0.2,0.2,0.75}

   \usepackage[normalem]{ulem}
   \usepackage{bm}

\makeatletter
\DeclareRobustCommand\onedot{\futurelet\@let@token\@onedot}
\def\@onedot{\ifx\@let@token.\else.\null\fi\xspace}

\makeatother

\def\clap#1{\hbox to 0pt{\hss #1\hss}}%
\def\initials#1{\protect\clap{\protect\smash{\protect\raisebox{1.4ex}{\protect\tiny{\protect\textsf{\protect\textit{#1}}}}}}}%
\makeatletter
\newcommand{\EDIT}[4][]{\protect\@ifundefined{hidecomments}{%
  \protect\strut{\color{#3}{\hspace{0pt}\initials{#2}\protect\sout{#1}{~#4}}}%
  }{#4}}
\newcommand{\NOTEboxed}[3]{\protect\@ifundefined{hidecomments}{%
  {\begin{center}\fbox{\parbox{0.97\linewidth}{\protect\EDIT{#1}{#2}{#3}}}\end{center}}
  }{}}
\newcommand{\COMM}[3]{\protect\@ifundefined{hidecomments}{%
  {\protect\EDIT{#1}{#2}{#3}}%
  }{}}
\newcommand{\DefAuthor}[2] 
{%
  \expandafter\newcommand\csname #1edit\endcsname[2][]{\protect\EDIT[##1]{#1}{#2}{##2}}
  \expandafter\newcommand\csname #1\endcsname[1]{\protect\COMM{#1}{#2}{[##1]}}
  \expandafter\newcommand\csname #1boxed\endcsname[1]{\protect\NOTEboxed{#1}{#2}{##1}}
}
\definecolor{dfltgreen}       {rgb}{0.0,0.5,0.0}
\definecolor{dfltred}         {rgb}{0.7,0.0,0.0}
\newcommand{\REVadd}[1]{\protect\@ifundefined{hidecomments}{%
  \strut{\color{dfltgreen}{#1}}}{#1}}
\newcommand{\REVedit}[2][]{\protect\@ifundefined{hidecomments}{%
  \strut{\color{dfltred}{\protect\sout{#1}}\color{dfltgreen}{~#2}}}%
  {#2}}
\makeatother

\definecolor{dkgreen}       {rgb}{0.0,0.5,0.0}
\definecolor{dkblue}        {rgb}{0.0,0.0,0.7}
\definecolor{dkcyan}        {rgb}{0.0,0.5,0.5}
\definecolor{dkmagenta}     {rgb}{0.5,0.0,0.5}
\DefAuthor{MK}{dkmagenta} 
\DefAuthor{MM}{dkgreen} 
\DefAuthor{VG}{dkblue} 
\DefAuthor{MF}{dkcyan} 
\DefAuthor{SA}{orange} 
\DefAuthor{todo}{red}

\usepackage{pifont}

\begin{document}
\title{{\method: Towards Learnable Quantization for Low-bit Pattern Recognition}}

\author{
Mishal Fatima\inst{1}$^\ast$ \and
Shashank Agnihotri\inst{1}$^\ast$ \and
Marius Bock\inst{2}$^\ast$ \and
Kanchana Vaishnavi Gandikota\inst{2} \and
Kristof Van Laerhoven\inst{2} \and
Michael Moeller\inst{2} \and
Margret Keuper\inst{1,3}
}
\renewcommand{\thefootnote}{}
\footnotetext{$^\ast$Equal contribution.}

\authorrunning{M. Fatima et al.}

\institute{University of Mannheim \\\and
University of Siegen \\ \and
MPI for Informatics, Saarland Informatics Campus}

\maketitle  

\begin{abstract}
Most pattern recognition models are developed on pre-proce\-ssed data. In computer vision, for instance, RGB images processed through image signal processing (ISP) pipelines designed to cater to human perception are the most frequent input to image analysis networks. However, many modern vision tasks operate without a human in the loop, raising the question of whether such pre-processing is optimal for automated analysis. Similarly, human activity recognition (HAR) on body-worn sensor data commonly takes normalized floating-point data arising from a high-bit analog-to-digital converter (ADC) as an input, despite such an approach being highly inefficient in terms of data transmission, significantly affecting the battery life of wearable devices. In this work, we target low-bandwidth and energy-constrained settings where sensors are limited to low-bit-depth capture. 
We propose \method, i.e.~the task-specific \textit{learning} of a non-linear quantization for pattern recognition. We exemplify our approach on raw-image object detection as well as HAR of wearable data, and demonstrate that raw data with a learnable quantization using as few as 4-bits can perform on par with the use of raw 12-bit data. All code to reproduce our experiments is publicly available via \url{github.com/Mishalfatima/Gamma-Quant}.
\end{abstract}

\section{Introduction}\label{sec:introduction}

Deep learning techniques have revolutionized the performance of numerous pattern recognition tasks in the last decade by training on large-scale image datasets. Yet, comparably little attention has been paid to the type and quantization of the input data. In computer vision, for instance, most pipelines consider pre-processed sRGB images with a standard bit depth of 8 bits. In the imaging process, cameras usually capture visual information in a higher-bit-depth RAW format which is converted to the standard format by an image signal processor (ISP) using a series of operations including black light subtraction, demosaicking, denoising, white balancing, gamma correction, color manipulation, and tone-mapping to finally obtain a visually pleasing 8-bit sRGB image. As photography-oriented ISP pipelines may not be optimal for vision tasks, recent works \cite{mosleh2020hardware,diamond2021dirty,robidoux2021end,Onzon_2021_CVPR}  have also successfully optimized ISP pipelines together with the downstream vision task. Yet, the idea to explicitly optimize the quantization for a given (automated) machine learning task has not been exploited so far. 

Similarly, most studies in human activity recognition (HAR) from body-worn data simply use linearly quantized high-bit (e.g.~12-bit) information from the sensors. While settings of lower bit quantizations have been investigated (see \cite{log_quant_Buckler_2017_ICCV,9191352}), the idea to \textit{learn} an optimal quantization has not been studied. 

In both settings, computer vision and HAR, the quantization of the analog data into a digital signal plays a critical role in balancing data quality, memory requirements, and energy consumption at the analog-to-digital converter (ADC), see e.g.~\cite{lenero2014review}. Moreover, significant bandwidth can be saved if the recorded data is sent to the cloud in a low-bit format for further analysis. 

In this paper, we demonstrate that a tailored \textit{learned} quantization has significant advantages over a na\"ive linear quantization. Specifically, we study a learnable non-linear quantization via
\begin{equation}
\label{eqn:our_gamma_teaser}
\mathcal{Q}(\mathcal{X}, \gamma, \mu) = Q_{\hat{N}}(\text{sign}(\mathcal{X}-\mu)\cdot|\mathcal{X}-\mu|^{\gamma})
\end{equation}
for (normalized) analog input values $\mathcal{X}$, a linear quantizer $Q_{\hat{N}}$ to a target bit-depth of $\hat{N}$ bit, and learnable parameters $\gamma$ as well as an offset $\mu$. This learnable non-linear quantization is optimized together with neural networks for specific tasks. 
We refer to our approach \eqref{eqn:our_gamma_teaser} as $\gamma$-QUANT.

We demonstrate that \method{} improves the performance of object detection on raw data on diverse vision datasets like the PASCAL-RAW dataset~\cite{omid2014pascalraw} and the RAOD dataset~\cite{xu2023RAODdataset} by simulating $\gamma$-QUANT on analog signals by using raw (high bit depths) images of the respective datasets. We show generality of the approach by conducting a similar study for a completely different modality, i.e., inertial, body-worn sensor data, using different datasets commonly used in HAR.

In summary, the contributions of this work are as follows:
\begin{itemize}
\item We show that na\"ive low-bit quantization of accelerometer data as well as of images harms model performance.  
\item We propose $\gamma$-QUANT, a learnable non-linear quantization (parameterized similar to a gamma correction), 
as a solution. 
\item We demonstrate that our proposed method allows reducing the recorded data to up to 4-bit for object detection and even 2-bit for human activity recognition without significant performance drops in comparison to high-bit data. Moreover, we demonstrate that learning the quantization via \method{} yields systematic improvements over a classical (linear) quantization. 
\end{itemize}

\section{Related Work}
\label{sec:related}
Work in the direction of jointly optimizing sensor and neural network parameters is limited, though some substantial contributions have been made recently. We summarize them in our two application domains, computer vision and wearable sensor data analysis, separately. 

\subsection{Quantization of Wearable Sensor Data}
Energy efficiency is crucial for wearable human activity recognition on edge devices with limited battery capacity. Transmission of HAR data accounts for significant energy consumption at the wearable device \cite{lara2012survey} which can be minimized by reducing expensive communications, for instance, by  aggregating or compressing data, and doing on-device feature extraction and classification to avoid sending raw signals. Recent work has focused on efficient inference of neural networks for on-device HAR using pruning \cite{liberis2023differentiable}, adaptive inference \cite{9669989} and quantization \cite{daghero2021ultra,daghero2022human,zhou2025efficient}. Unlike the quantization of HAR data, network quantization has been widely studied. Exemplary approaches include sub-byte and mixed-precision quantization with adaptive inference in 1D CNNs \cite{daghero2022human}, full-integer quantization of DeepConv LSTM \cite{zhou2025efficient}, and binary quantization of weights and activations in neural networks \cite{daghero2021ultra}.  Orthogonal to these techniques, power savings can also be achieved by turning off the sensors when inactive  or lowering their sampling rates \cite{5665860,zheng2017novel} or adapting sampling rates per activity \cite{6246136,cheng2018learning}. 
Further techniques have also been proposed to handle such HAR data captured at variable sampling rates, including modifications to neural network architectures \cite{malekzadeh2021dana}, and data augmentation \cite{9260208}. Unlike these techniques, we address the often-overlooked challenge of reducing energy use during data capture by applying low-bit quantization directly at the inertial sensor, which can provide task-aware data compression immediately at acquisition, complementing the existing energy-saving techniques.

\subsection{Codesigning Imaging System and Vision Models}
Instead of the traditional approach where imaging hardware and perception models are developed independently, a recent trend in efficient machine learning is to code-sign imaging systems and computer vision models \cite{klinghoffer2022physics}, creating tightly integrated solutions that maximize performance while reducing the hardware or computing requirements. \cite{klinghoffer2023diser} formulate imaging building blocks as context-free grammar whose parameters can be optimized through a reinforcement learning framework. \cite{fu2021sacod,chang2019deep,chang2018hybrid} jointly optimize for the downstream computer vision model along with the optical layer to exploit its
potential computation capability.
\cite{sommerhoff2023differentiable} proposes a differentiable approach to jointly optimize the size and distribution of pixels on the imaging sensor along 
with the downstream computer vision model. 
However, none of these works deal with quantization at the sensor, which is the focus of our work.
\vspace{0.1cm}

RAW images contain more information than standard RGB images (sRGB), which could potentially be beneficial for higher-level vision tasks such as object detection \cite{ljungbergh2023raw,xu2023RAODdataset,wu2024dense}. In this context, a hardware-in-the-loop method is introduced in \cite{mosleh2020hardware} to optimize hardware ISP for end-to-end task-specific networks by using zero-order optimization. 
Diamond \emph{et al.} \cite{diamond2021dirty} use Anscombe networks as neural ISPs which are jointly trained with task-specific neural networks. \cite{morawski2022genisp}  train a minimal neural ISP pipeline for object detection that improves
generalization to unseen camera sensors.  Robidoux \emph{et al.}\cite{robidoux2021end} optimize HDR ISP hyperparameters together with detector network, whereas \cite{Onzon_2021_CVPR} train a neural network for automatic exposure selection that is trained jointly with ISP pipeline and a network for HDR object detection.
Instead of using the whole ISP, \cite{log_quant_Buckler_2017_ICCV,xu2023RAODdataset} identifies the key components of the ISP for downstream vision tasks. 
Prior works \cite{ljungbergh2023raw,xu2023RAODdataset,liu2024learnable} learn specific parameterized ISP functions such as demosaicking and gamma correction, color correction in an end-to-end fashion along with the object detector. While some of these works  \cite{ljungbergh2023raw,xu2023RAODdataset} learn a gamma correction, this is a part of the ISP pipeline after a digital image has been obtained, while we propose to learn this function before quantization at the sensor for the analog input. 
While \cite{log_quant_Buckler_2017_ICCV,9191352} consider the effect of low bit quantization on vision tasks, they do not optimize this process.
Instead of learning fixed ISP parameters, \cite{NEURIPS2024_cc596a80} introduces a scene-adaptive ISP to automatically generate an optimal ISP pipeline and the corresponding ISP parameters to maximize the detection performance. 
Yet, the learned computational steps are applied to the digital (= already quantized) image. To the best of our knowledge the \textit{learning} of a quantization to be applied within the analog-to-digital converter of a camera, has not been considered before. 

\section{Our Approach \method}
\label{sec:method}
\subsection{Preliminaries}
\label{sec:preliminaries}

Many sensors use a physical effect to induce a voltage which is converted to a digital measurement by an analog-to-digital converter (ADC). For instance, variable capacitance Micro-Electro-Mechanical Systems (MEMS) accelerometers that are present in many mobile and wearable systems, sense tiny mass' distance changes between two capacitor plates. An in-chip ADC converts the resulting amplified voltages to quantized digital values that are handled in a local processing unit. Similarly, in imaging, a CMOS sensor uses photodiodes (and the photo-electric effect) to induce a charge, which is transferred to a capacitor. The charge is converted to a voltage, amplified and passed to the ADC. For both modalities the ADC typically uses a linear quantization with at least 8, more commonly 12 or even 16 bits per value. Yet, both the energy consumption as well as the readout speed are crucially influenced by the bit-depth, such that a reduction can have significant benefits: According to \cite{teledyne_bitdepth_speedup}, simply reducing the image bit-depth from 16-bit to 12-bit has shown speedups by a factor of two in industry. It was shown in \cite{chae20102} that -- for energy-constrained sensors -- ADCs alone contribute up to 50\% of the energy consumption in an image sensor. Similarly, in wearable sensorics, transferring the recorded accelerometer data wirelessly has shown to be the by far most energy-consuming operation, such that a reduced bit-depth has an immediate and significant effect on the wearable's battery life.

\subsection{A Learnable Quantization Approach}
The most straight-forward ADCs produce lower bit-depths via linear quantization, i.e., converting each analog value $\mathcal{X}$ (assumed to be normalized to $[0,1]$) to 
a digital value $\mathcal{X}_\mathrm{Q}$ via
\begin{equation}
    \label{eqn:linear_quantization}
    Q_{\hat{N}}(\mathcal{X}) = \left\lfloor{\mathcal{X}\cdot (2^\mathrm{\hat{N}}-1)}\right\rfloor ,
\end{equation}
where $\mathrm{\hat{N}}$ is the desired bit depth, and $\lfloor{.}\rfloor$ is the floor operation. 

\begin{figure}[t]
    \centering
    \includegraphics[width=1.0\linewidth]{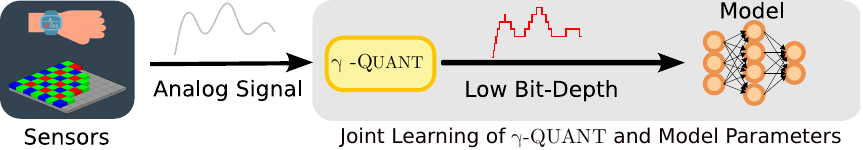}
    \caption{\method{} learns the quantization of an ADC to convert the analog signal of a sensor to a low bit depth digital signal, which is subsequently sent to the neural network for performing the downstream task. The parameters of the quantization are trained jointly with the parameters of the neural network in a task-specific fashion. \method{} can reduce the energy consumption of the sensor significantly with a minimal loss in performance.}
    \label{fig:method}
\end{figure}

In imaging, the Image Signal Processor (ISP) converts the raw digital values to standard RGB (sRGB) images through a sequence of pre-processing steps including demosaicking, denoising, white-balancing, color conversion, and tonemapping. Yet, low bit-depth raw-images often lead to poor visual quality. As the human visual system rather scales logarithmically, 
\cite{log_quant_Buckler_2017_ICCV,adaptive_log_pixel_sensor} proposed to scale the analog signals before quantization via
\begin{equation}
 \mathcal{X}_{\mathrm{log}} = \log(\mathcal{X}+\epsilon) ,
\label{eq:log_scale}
\end{equation}
where $\epsilon$ is required to bound the input to the logarithm from below. \cite{log_quant_Buckler_2017_ICCV} use $\epsilon$=1 and quantize the resulting signal via
\begin{equation}
 \mathcal{X}_{\mathrm{Q}} = \left\lfloor{ \frac{\mathcal{X}_{\mathrm{log}} - \mathrm{min}(\mathcal{X}_{\mathrm{log}})}{\mathrm{max}(\mathcal{X}_{\mathrm{log}}) - \mathrm{min}(\mathcal{X}_{\mathrm{log}})} \cdot (2^\mathrm{\hat{N}}-1)}\right\rfloor.
\label{eq:log_quant}
\end{equation}
Yet, the quantization is ad-hoc, independent of the target bit-depth $\hat{N}$, and the effect of $\epsilon$ heavily depends on the dynamic range of the analog signal, such that it might need domain- and downstream-task specific adaption.

For body-worn accelerometer data, values are typically in $[-1,1]$ (up to a scaling) as accelerations for each axis can happen in two (opposite) directions, reflected by different signs. Thus, a logarithmic quantization is not straightforward. Moreover, accelerometers might constantly yield a non-zero signal in the gravitational field of the earth such that zero might not be a natural candidate for the finest quantization anymore. 

For these reasons, we propose \method{}, i.e., an automated approach for \textit{learning} the quantization of sensor signals in an automated and \textit{task-specific} way, see Fig.~\ref{fig:method}. More precisely, we propose to learn a non-linear low-bit ADC parameterized via \eqref{eqn:our_gamma_teaser} along with a neural network for a specific task by optimizing
\begin{equation}
 \min_{\theta, \gamma, \mu} ~\mathbb{E}_{(\mathcal{X},\mathrm{y})}\left[\mathcal{L}(\eta(\mathcal{Q}(\mathcal{X}, \gamma, \mu);\theta),\mathrm{y})\right],
     \label{eq:joint_learning}
 \end{equation}
  where $\eta$ is the task-specific neural network with parameters $\theta$, $\gamma$ and $\mu$ are the parameters of the learnable ADC (i.e., the quantizer $\mathcal{Q}$), and $\mathcal{L}$ is a suitable loss function comparing the network output and the ground truth prediction $\mathrm{y}$. 

To optimize \eqref{eq:joint_learning}, we propose to \textit{simulate} the learnable ADC $\mathcal{Q}(\mathcal{X}, \gamma, \mu)$ by using readily available high-bit data $\hat{\mathcal{X}}$ as an approximation to the analog signal $\mathcal{X}$. Since the quantization operation stops the flow of gradients in the backward pass, we use a straight-through estimator to allow gradient-based optimization of $\gamma$ and $\mu$. 

While a simulation enables the training of \eqref{eq:joint_learning} with standard first-order methods on large data sets, the resulting learned (non-linear) ADCs can be realized in hardware: For body-worn sensors such as 3D accelerometer MEMS, the in-chip Digital Signal Processor can be re-configured to adjust ADC settings such as the resolution or sensitivity range through internal registers. These subsequently alter the step size or quantization level of the digital output. Custom transfer functions or lookup tables can be implemented in the sensor to apply non-linear scaling. For CMOS image sensors non-linear quantization in the ADC can also be realized as demonstrated in  \cite{2006AnalogGammaCorrection,2021AnalogGammaCorrection}. Thus, our learnable quantization framework targets specific application (such as smart watches for human activity recognition or object detection cameras in an autonomous driving setting) where dedicated hardware is built/programmed using the task-specific learned quantization. 

\subsection{Specifics of \method{} for HAR and Object Detection}
\begin{figure*}[h]
    \centering
    \includegraphics[width=1.0\linewidth, trim={0cm 18cm 0cm 0cm}, clip]{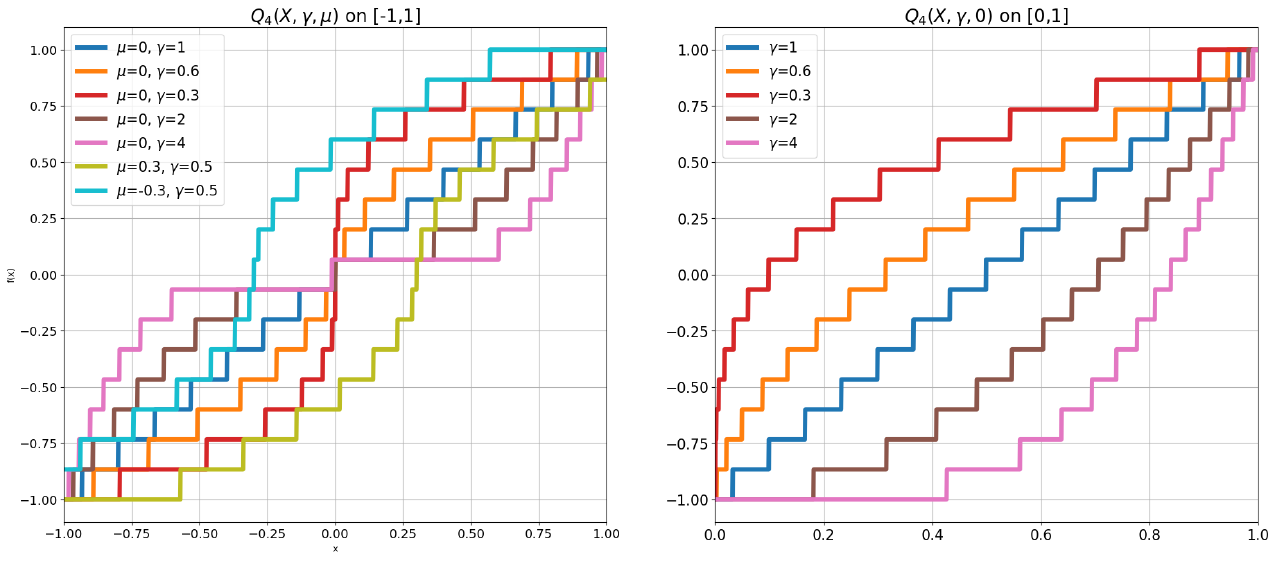}
    \caption{Exemplifying different \method{} quantization for HAR on $[-1,1]$ data (left) and object detection on $[0,1]$ data (right).}
    \label{fig:quantization_curves}
\end{figure*}

When applying \method{} to imaging applications, we built upon findings of priors works (e.g. \cite{log_quant_Buckler_2017_ICCV}) that low intensity values are important to resolve in a more fine-grained manner. Therefore fix the offset $\mu=0$, and simplify \method{} to 
\begin{equation}
\label{eqn:our_gamma}
\mathcal{Q}(\mathcal{X}, \gamma) = Q_{\hat{N}}(\mathcal{X}^{\gamma}) =  \left \lfloor \mathcal{X}^{\gamma} \cdot (2^{\hat{N}-1}) \right \rfloor,
\end{equation}
where we assume $\mathcal{X}$ to be normalized to $[0,1]$. More specifically, when simulating $\mathcal{X}$ from a high-bit digital signal, we divide the digital signal by $2^N -1$ for an $N$-bit signal. For a faithful simulation of an analog signal, $N$ needs to be significantly larger than our target bit depth $\hat{N}$. The resulting \method{} approach can learn to resemble a simple linear quantization ($\gamma = 1$), compress large values more strongly ($\gamma<1$), or compress small values more strongly ($\gamma>1$), see Fig.~\ref{fig:quantization_curves}, right. 

When applying \method{} to body-worn sensors for HAR, we use the full approach \eqref{eqn:our_gamma_teaser} as data is typically normalized to $[-1,1]$ and offsets can be important. Explicitly writing out the linear quantizer for $[-1,1]$ our approach becomes
\begin{equation}
\label{eqn:our_gamma_har}
\mathcal{Q}(\mathcal{X}, \gamma, \mu) = \text{round}\left(((\text{sign}(\mathcal{X}-\mu)\cdot|\mathcal{X}-\mu|^{\gamma})+1)/2 \cdot (2^N-1)\right),
\end{equation}
resulting in exemplary curves illustrated in Fig.~\ref{fig:quantization_curves} on the left. As the function \eqref{eqn:our_gamma_har} is non-differentiable at $\mathcal{X}=\mu$ for $\gamma<1$, we found the replacement of $|\mathcal{X}-\mu|^{\gamma}$ by $(|\mathcal{X}-\mu|+\epsilon)^{\gamma}$ for a small $\epsilon$, e.g. $\epsilon=10^{-3}$ to stabilize the training process. 

\section{Experiments on Learnable Quantizations for Accelerometer-based Human Activity Recognition}
\label{sec:inertialexperiments}
As a first use case of our \method{} we focus on inertial sensors, specifically accelerometers used in the context of Human Activity Recognition. The following will outline datasets used as well as experiments conducted. As accelerations measured by wearable sensors can go in both directions, value ranges of the investigated datasets are between $[-1, 1]$. We thus apply the version of \method{} for wearable accelerometer data as shown in \eqref{eqn:our_gamma_har}. 

\subsection{Datasets}
In total, we investigate five wearable accelerometer datasets (see Table~\ref{tab:inertialdatasets}). The WEAR dataset \cite{bockWEAROutdoorSports2024} records participants outdoors, while performing a set of workout-related activities such as running, stretching and strength-based activities. Similarly, the Hang-Time \cite{hoelzemannHangTimeHARBenchmark2023} dataset records a team of basketball players during their practice session consisting of a warm-up, drill and game session. Recorded in a biology wet lab, the Wetlab dataset \cite{schollWearablesWetLab2015} records recurring activities, such as pipetting, occurring during a DNA extraction experiment. Lastly, the RWHAR \cite{sztylerOnBodyLocalizationWearable2016} and SBHAR dataset \cite{reyes-ortizTransitionAwareHumanActivity2016} has participants consists of various of locomotion activities such as walking stairs, with the SBHAR dataset providing annotations of additional transitional activity periods that mark the transition from one to another activity. Note that all but the RWHAR dataset provide continuous recording data, thus providing an additional \textit{NULL}-class which represents times during the recording participants did not perform any of the activities of relevancy.

\begin{table}
\scriptsize
\centering
\setlength{\tabcolsep}{0.2em}
\caption{Investigated HAR datasets. Table provides: participant count, activity count (classes), sensor axes count and overall recording scenario. Each sensor axes provides accelerometer data sampled at 50Hz.}
\label{tab:inertialdatasets}
{\renewcommand{\arraystretch}{1.2}
\begin{tabular}{lccccl}
 Dataset        & participants & classes & axes & scenario \\ \toprule
 WEAR \cite{bockWEAROutdoorSports2024}                      & 18 & 19 & 12 & body-weight workout \\
 Wetlab \cite{schollWearablesWetLab2015}                    & 22 & 9  & 3 & laboratory \\
 Hang-Time \cite{hoelzemannHangTimeHARBenchmark2023}        & 24 & 6  & 3 & basketball \\
 RWHAR \cite{sztylerOnBodyLocalizationWearable2016}         & 15 & 8  & 21 & locomotion \\
 SBHAR \cite{reyes-ortizTransitionAwareHumanActivity2016}   & 30 & 13 & 3 & locomotion + transitional \\ \bottomrule
\end{tabular}
}
\end{table}

\subsection{Experimental Setup and Implementation}  
\label{sec:inertialexp}
During experiments, we follow a Leave-One-Subject-Out (LOSO) cross-validation, where each participant in the dataset is used as the validation set exactly once, while all other participants are used for training. We report the class-averaged macro F1-score averaged across all validation splits (i.e., participants). We further seed-average our experiments repeating each experiment using a set of three different random seeds. For all experiments we employ a sliding window of one second with a 50\% overlap, normalize all accelerometer signals between $[-1,1]$ using min-max normalization applied across the full dataset, a weighted cross-entropy loss and the Adam optimizer, with a learning rate of $1e^{-4}$, weight decay of $1e^{-6}$. We train for 30 epochs with a batch size of 100, using a learning rate schedule that multiplies the learning rate by a factor of 0.9 every 10 epochs. As a model architecture of choice we use the DeepConvLSTM, a widely-adopted HAR model \cite{ordonezDeepConvolutionalLSTM2016}.
For each dataset we compare results using linear quantized and \method{} quantized accelerometer signals as input, differing bit depths to be either 2 or 4. We further provide results using the raw accelerometer signal as provided by the datasets. We further compare a dataset-wide versus sensor-axis-specific learnable quantization using \method{}, i.e., learning a $(\gamma, \mu)$-pair for each sensor axis in the dataset. In all experiments using \method{} we initialize $\gamma$ with 0.4, as we generally estimate differences accelerations close to $0$, i.e., fine-grained movements, to contain more information than the differences in large accelerations, i.e., strong movements. 

\subsection{Evaluation Results}  
\label{sec:inertialresults}
Table~\ref{tab:inertialresults} presents the per-dataset results of the experiments described above. It is evident that with 4-bit linear quantization, performance on the WEAR and Wetlab dataset drops significantly compared to using raw data. With 2-bit linear quantization, all five investigated HAR datasets witness a substantial decline in performance. However, across all datasets, \method{} consistently outperforms models trained on linearly quantized data. In particular, for the WEAR and Wetlab datasets, sensor-axis-specific learnable quantizations lead to notable improvements, achieving prediction F1-scores comparable to raw data, even with 2-bit input.

\begin{table}
\scriptsize
\centering
\caption{Per-dataset HAR results comparing training using raw data with training using linear, dataset-wide \method{} or per-axis \method{} quantized accelerometer signals. We provide results for bit depths $\hat{N}=2$ and $\hat{N}=4$.}
\setlength{\tabcolsep}{0.2em}
\label{tab:inertialresults}
{\renewcommand{\arraystretch}{1.2}
\begin{tabular}{lc c c c c c}
 &                                & WEAR            & Wetlab           & Hang-Time        & RWHAR           & SBHAR             \\ \toprule
 & \textit{Raw data}              & $\mathit{71.01 \pm0.18}$ & $\mathit{25.61 \pm 0.07}$ & $\mathit{33.88 \pm 0.09}$ & $\mathit{71.21 \pm0.90}$ & $\mathit{54.58 \pm 0.22}$  \\ \midrule
 \multirow{3}{*}{\rotatebox[origin=c]{90}{$\hat{N} = 4$}}
 & linear                         & $63.74 \pm0.22$ & $21.72 \pm0.18$ & $33.69 \pm0.22$ & $69.78 \pm0.63$ & $52.14 \pm0.26$  \\
 & \method                        & $70.14 \pm0.51$ & $23.00 \pm0.24$ & $33.63 \pm0.17$ & $69.28 \pm0.46$ & $\mathbf{52.58 \pm0.20}$  \\
 & \method (per-axis)             & $\mathbf{70.17 \pm0.39}$ & $\mathbf{23.63 \pm0.20}$ & $\mathbf{33.67 \pm0.17}$ & $\mathbf{71.16 \pm1.26}$ & $52.52 \pm0.21$  \\ \midrule
 \multirow{3}{*}{\rotatebox[origin=c]{90}{$\hat{N} = 2$}}
 & linear                         & $58.91 \pm0.13$ & $12.62 \pm0.34$ & $28.01 \pm0.14$ & $64.82 \pm0.62$ & $33.02 \pm0.39$  \\
 & \method                        & $60.45 \pm0.43$ & $12.94 \pm0.18$ & $30.67 \pm0.12$ & $\mathbf{71.41 \pm0.51}$ & $40.18 \pm0.32$  \\
 & \method (per-axis)             & $\mathbf{63.95 \pm0.14}$ & $\mathbf{16.85 \pm0.54}$ & $\mathbf{30.60 \pm0.22}$ & $69.21 \pm0.32$ & $\mathbf{41.88 \pm0.32}$ \\ \bottomrule
\end{tabular}
}
\end{table}

\begin{figure}%
    \centering
    \begin{minipage}{.5\textwidth}
        \centering
        \includegraphics[width=0.9\linewidth, trim={2cm 2cm 2cm 2cm}, clip]{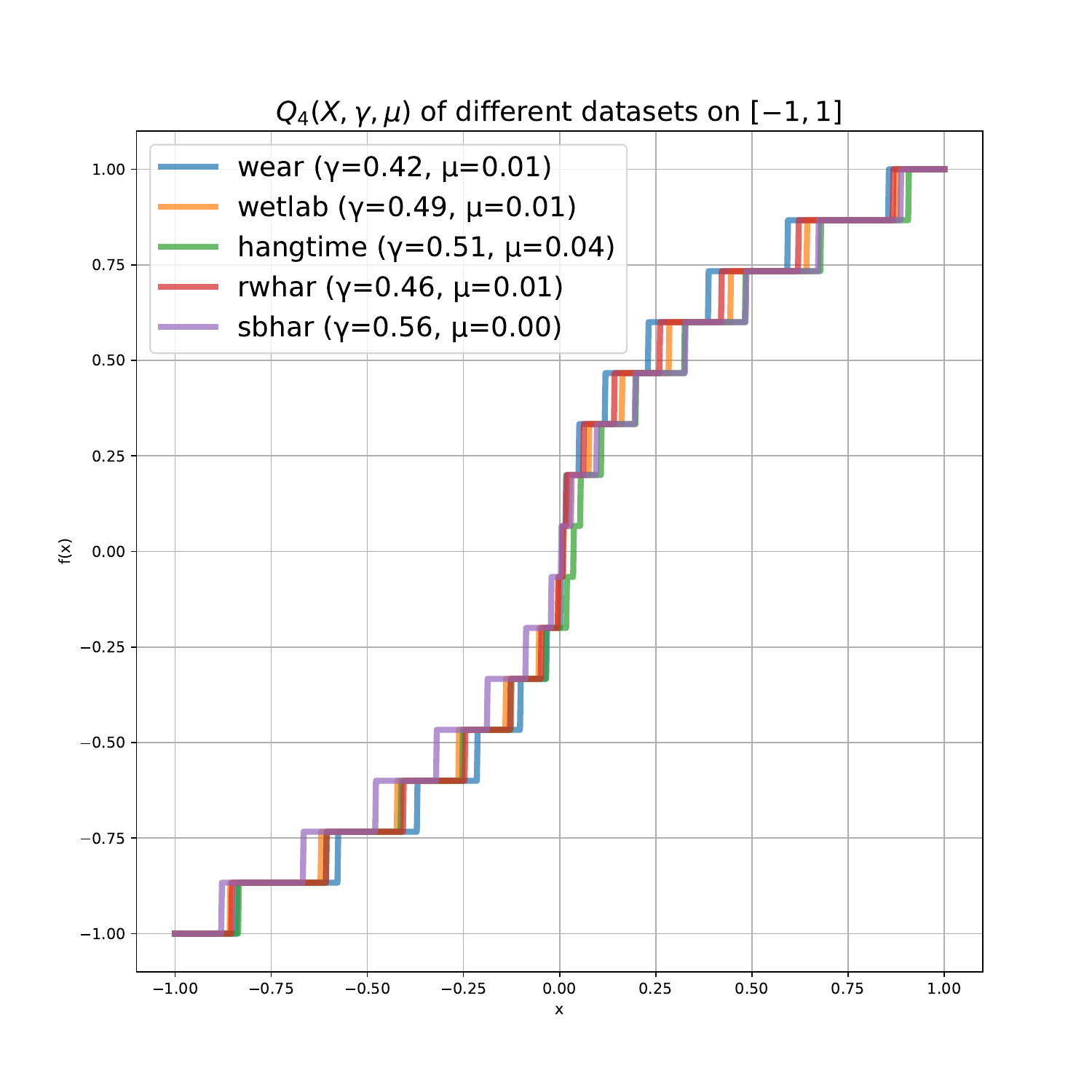}
    \end{minipage}%
    \begin{minipage}{.5\textwidth}
        \centering
        \includegraphics[width=0.9\linewidth, trim={2cm 2cm 2cm 2cm}, clip]{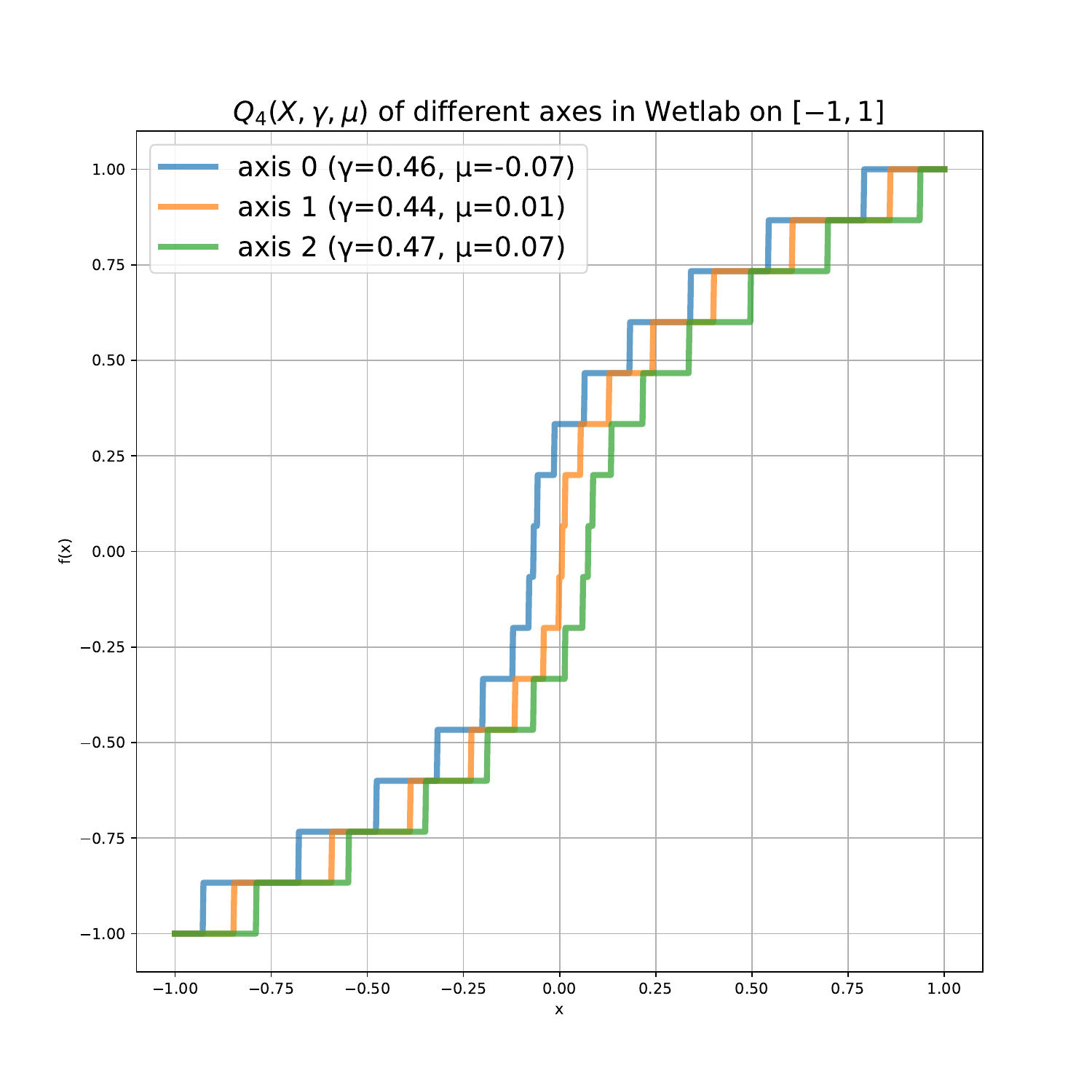}
    \end{minipage}
    \caption{Visualization of exemplary learned quantization function using \method{}. The left plot shows the per-dataset learned quantization functions. The right plot shows the per-axis quantization functions when predicting the Wetlab dataset. $\gamma$ and $\mu$ values in both plots are averaged across validation splits.}
    \label{fig:inertialgammafunctions}%
\end{figure}

Figure~\ref{fig:inertialgammafunctions} illustrates representative quantization functions learned during HAR experiments. The left subplot shows that different datasets yield distinct global scaling factors $\gamma$, while the learnable offset $\mu$ remains close to zero across datasets. However, as demonstrated in the right subplot for the Wetlab dataset, if \method{} is learned individually on a per-axis level, both a negative offset ($\mu<0$) for the x-axis as well as a positive offset ($\mu>0$) for the z-axis of the sensor are learned. Supported by the overall pattern of improved prediction results of the per-sensor application of \method{}, we hypothesize that the learned offsets allow a better compensation of the static gravitational components captured by fixed accelerometers, where certain axes exhibit constant acceleration depending on their orientation relative to the ground. This compensation works especially well for sensors whose orientation is rather static, e.g., because they are placed at body parts that do not exhibit strong rotations, e.g., the chest or waist of a participant, or capture mostly activities, which do not exhibit strong movements such as sitting. Conversely, one can only expect smaller gains from per-axis quantization on sensors which exhibit frequent changes in orientation due to fast movements of e.g. sport-related activities such as present in the Hang-Time dataset, as gravitational components do not remain static.

\section{Experiments on Learnable Quantizations for Object Detection on CMOS Sensor Data}
\label{sec:experiments}

\begin{figure}[h]
    \centering
    \includegraphics[width=1.0\linewidth]{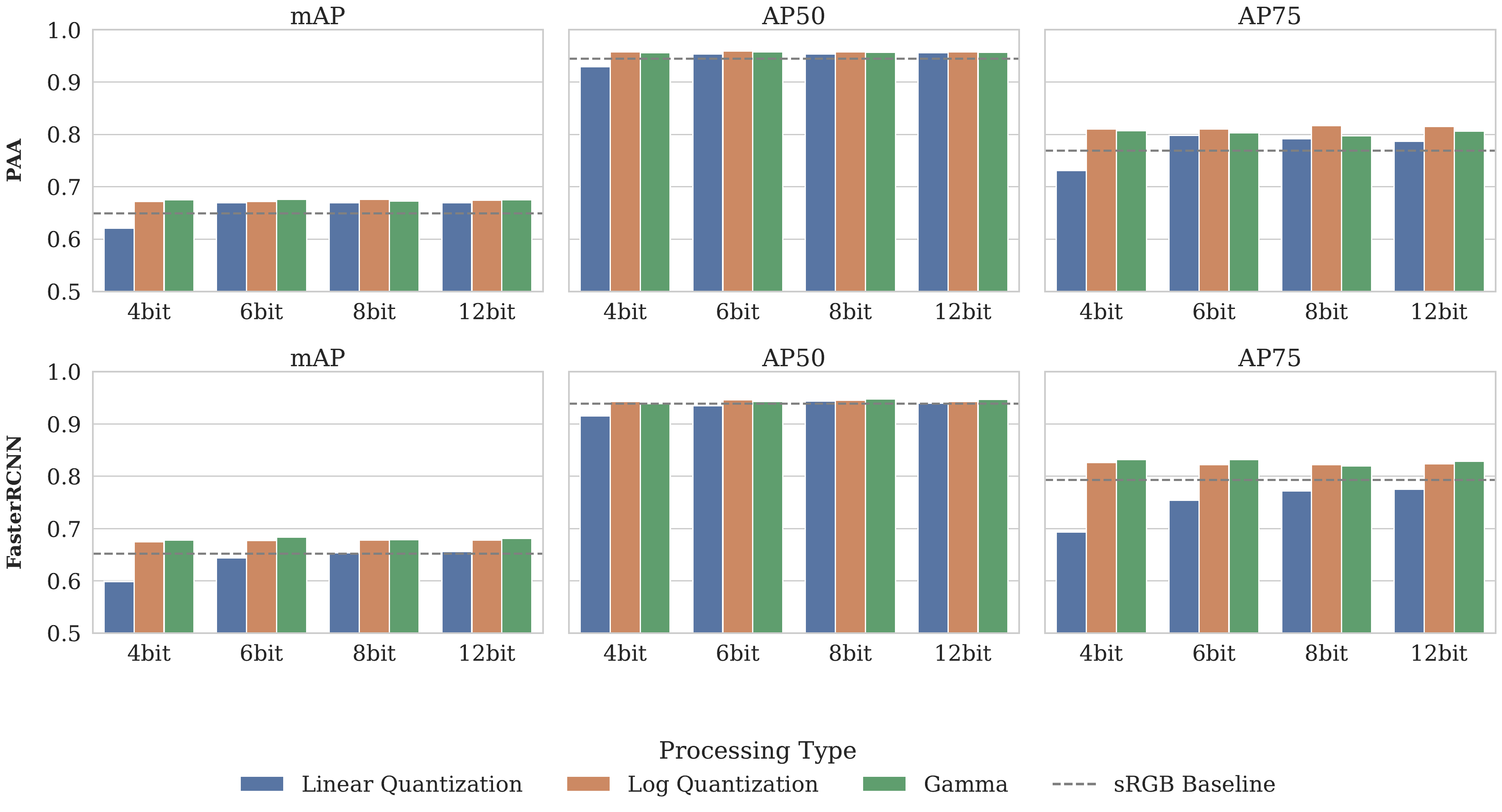}
    \caption{Results with \method{}, log and linear quantization on PAA and FasterRCNN for different bit depths. 
     The blue dashed lines represented the performance of the model on sRGB input images.}
    \label{fig:all_mAP_rsults}
\end{figure}

\subsection{Datasets}
We use the PASCAL RAW dataset~\cite{omid2014pascalraw} having a bit-depth of 12, with three classes, namely `Car', `Bicycle', and `Person'. 
The training data consists of 2129 images, and the test data is 2130 images. Experiments on a second data set, RAOD~\cite{xu2023RAODdataset}, can be found in the appendix.

\subsection{Experimental Setup and Implementation}  
\label{sec:exp}
We use the mmdetection~\cite{chen2019mmdetection} framework from openmmlab for conducting experiments. 
Following previous works~\cite{xu2023RAODdataset,omid2014pascalraw} that perform object detection on RAW images, we used  Faster-RCNN~\cite{faster_rcnn}, and PAA~\cite{kim2020probabilistic_paa} models with pre-trained ResNet50~\cite{he2016deep} backbone. 
We conduct experiments by quantizing the Bayer patterned higher bit-depth RAW images to lower bit depths of 4, 6, 8, and 12 using \method{} and compare against Linear quantization (Linear Quant) as well as the logarithmic quantization \eqref{eq:log_quant}. The Bayer pattern RGGB image is viewed as a four-channel image, subsampled, and converted to a three-channel RGB image by averaging the two green channels. All the images are scaled to a range of [0,255] before being used as input to the neural network. 
  We report the following quantitative evaluation metrics: mean average precision (mAP) is calculated by averaging the Average Precision (AP) values across the intersection over union (IoU) thresholds ranging from [0.5, 0.95] with a step size of 0.05, resulting in 10 threshold values. 
We also report performance with IoU-thresholds of 0.5 and 0.75 (AP50 and AP75) for each case. 

Following the multi-scale training setup commonly used~\cite{faster_rcnn,lin2017focal}, during training, the shorter image side is scaled to one of the sides randomly selected from a set of sides: [480, 512, 544, 576, 608, 640, 672, 704, 736, 768, 800] using nearest neighbor interpolation and the longer side is scaled to maintain the aspect ratio. 
At test time, the shorter edge length is kept at 800. 
We use a batch size of 16 and train the FasterRCNN model for 140 epochs and PAA for 70 epochs. 
For FasterRCNN, we apply warm-up for the first 1000 iterations, linearly increasing the lr from  $1e^{-4}$ to $2.5e^{-3}$ followed by a multi-step learning rate scheduler. 
We use SGD with Nestrov Momentum~\cite{sgd_nestrov} as the optimizer with a weight decay of $1e^{-4}$. For PAA, we apply a warmup strategy for the first 4000 iterations, followed by cosine annealing for the learning rate schedule. The base learning rate is set to $1e^{-3}$ , with a weight decay of $1e^{-3}$.

\subsection{Evaluation Results}  
\label{sec:results}
To demonstrate the effectiveness of the proposed \method{}, we design a set of experiments across 4 different quantization levels, i.e., 4, 6, 8, 12 bits. 
For every quantization level, we run three quantization methods, namely, linear quantization, logarithmic quantization as shown in \eqref{eq:log_quant}, and \method{}. 
The quantitative results obtained for different bit depths using the three models on PASCAL RAW are presented in  \cref{fig:all_mAP_rsults}. 

We observe that performance with \emph{4-bit linear quantization is the worst} across all architectures. 
Low-bit linear quantization is especially hurting details in lower luminance regions, which account for a majority of the intensity values and therefore result in the lowest performance. Results with \method{} show systematic improvement across both architectures. In particular, there is surprisingly little difference between the different bit depths, indicating that - although a 4-bit image might not be visually pleasing - it is a sufficient bit depth for faithful object detection. Moreover, it is highly encouraging that the performance on standard (ISP processed) RGB images is met or even surpassed: \method{} learns to scale the distribution of pixel values such that low-intensity pixels are amplified, thus providing more contrast in the input.
This allows features and details to be better visible, enabling the model to learn more informed feature representations. 

Our experiments demonstrate that the log-quantization \eqref{eq:log_quant} performs on par with our proposed \method{}. While this might be discouraging at first glance, such results are based on a well chosen value of $\epsilon$, i.e., $\epsilon=1$ when simulating analog signals with digital 12-bit raw data or, correspondingly, $\epsilon\approx 0.00024414$ for analog signals scaled to $[0,1]$. More concretely, for an analog signal, there is no natural $\epsilon$, such that it becomes a hyperparameter to be tuned. Our framework can be seen as an automatic (differentiable) way of learning such a hyperparameter. In particular, $\epsilon$ in log-quantization could also be learned in the same framework. While this (and further much more flexible) parametrizations of the quantization are an interesting direction for future research, we decided to study \method{} for the sake of simplicity: The range of the output in the log-quantization depends on $\epsilon$ such that a rescaling to [0,1] needs to be included in the learning. Our \method{} automatically preserves the $[0,1]$ range. We thus consider the fact that \method{} reaches the performance of a logarithmic quantization to be encouraging. In particular, for FasterRCNN, the learned values of $\gamma$ are 0.294, 0.338, 0.354, 0.359 for 4, 6, 8, and 12 bits, respectively. Thus, the logarithmic shape of the curve is learned to be optimal and does not need to be derived from prior assumptions that might be violated in significantly different application scenarios. 

\section{Conclusion}
\label{sec:conclusion}
Our proposed task-specific quantization framework, \method{} offers a significant advancement in optimizing the non-linear quantization of ADCs for pattern recognition with different modalities. 
By moving beyond traditional high-bit-depth linear quantization and manual choice of a non-linear quantization (e.g., based on human perception), we develop an automatic task-aware quantization framework. 
This helps achieve substantial improvements in object detection and human activity recognition from body-worn sensors with sensor hardware constraints such as energy consumption and memory usage compared to traditional data processing pipelines. 
Our results highlight the potential of \method{} to maintain the performance of high-bit data in different tasks while minimizing energy consumption, memory usage, and data transmission costs, ultimately contributing to more efficient and sustainable machine learning workflows. While our current work is focused on hardware constraints at the sensor, in the future, we would expand on this framework to combine efficient network architectures and network quantization for inference, offering further improvements in computational efficiency.

\noindent\subsubsection*{Limitations: } It is impossible to have a dataset with truly analog signals.
Thus, in our work, high-bit depth RAW images serve as a proxy for analog signals. 
While this is a valid assumption that stems its roots in signal processing theory, there is still some loss of information between true analog and high-bit depth RAW input. Ideally, we would like to test our proposed \method{} on sensors directly to work on analog signals, with the potential benefit of further gains in accuracy.

\section*{Acknowledgments}
This work was supported by the DFG research project WASDEO (grant number 506589320) and by the DFG Research Unit 5336 – Learning to Sense (L2S). We further gratefully acknowledge the University of Siegen’s OMNI cluster and the high-performance computer HoreKa at the National High-Performance Computing Center at the Karlsruhe Institute of Technology (NHR@KIT) for providing computational resources.

\newpage
{
\newpage

\bibliographystyle{splncs04}
\bibliography{main}
}
\newpage
\appendix
\section*{Table Of Content}
The supplementary material covers the following information:
\begin{itemize}
\setlength\itemsep{2em}
\vspace{1em}
    \item \Cref{sec:appendix:visualizations}:  Visualizations
    \item \Cref{sec:appendix:method}: Additional Results: RAOD Dataset.
    \item \Cref{sec:appendix:gammahar}: Additional Results: Gamma Initialization (HAR).
\end{itemize}



\section{Visualizations}
\label{sec:appendix:visualizations}
\begin{figure*}
    \centering
    \resizebox{\linewidth}{!}{
    \begin{tabular}{@{}p{3mm}@{\hspace{1mm}}c@{\hspace{1mm}}c@{\hspace{1mm}}c@{\hspace{1mm}}c@{\hspace{1mm}}c@{\hspace{1mm}}c@{\hspace{1mm}}c@{}}
         & Log Quant ~\cite{log_quant_Buckler_2017_ICCV} & $\gamma\approx$Log Quant  & $\gamma$=0.1 Scale & $\gamma$=0.3 Scale & $\gamma$=0.5 Scale & $\gamma$=0.7 Scale & $\gamma$=0.9 Scale \\
         
         \rotatebox{90}{\tiny PASCAL RAW} & 
         \includegraphics[width=0.23\linewidth]{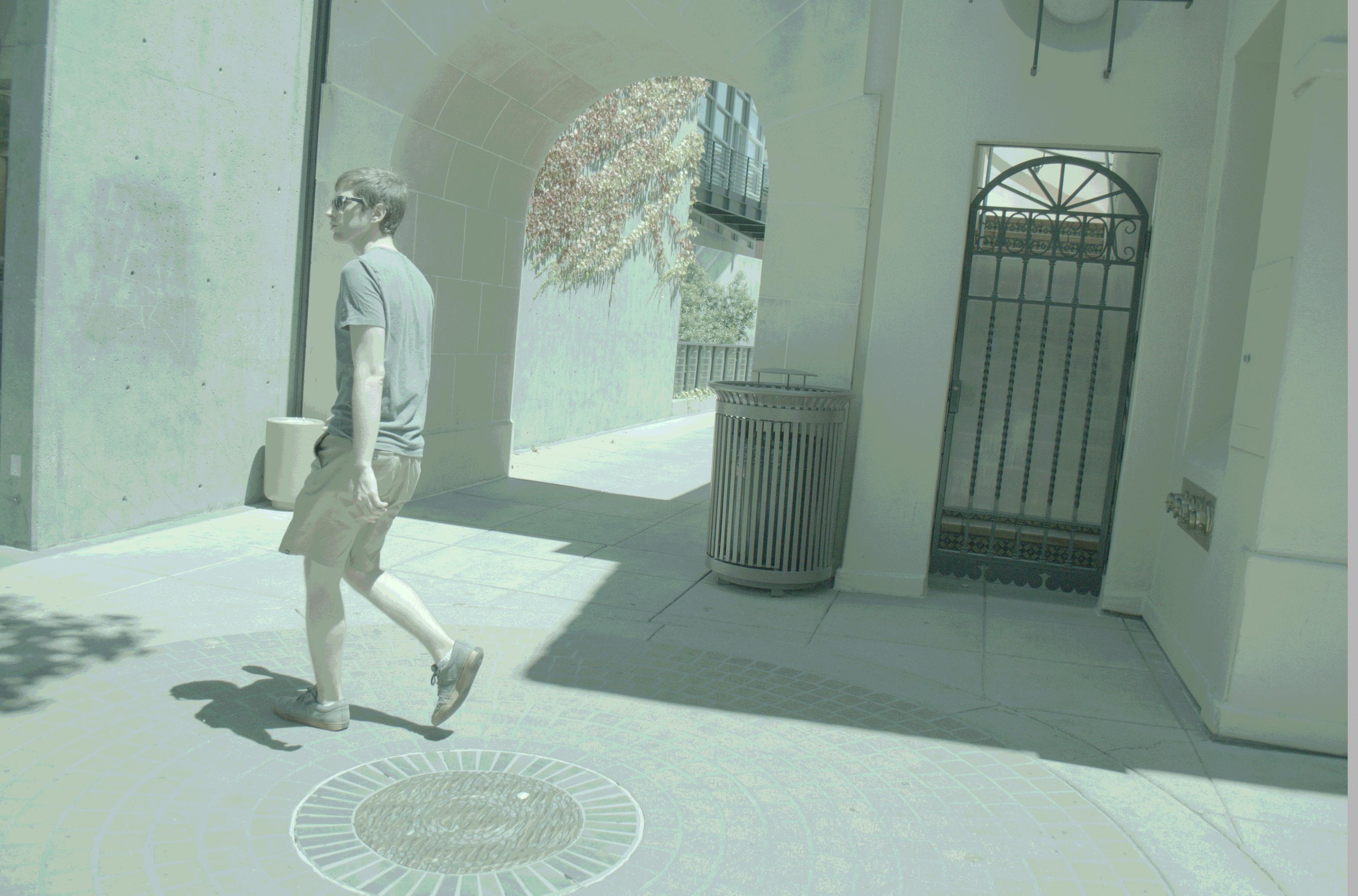}& 
         \includegraphics[width=0.23\linewidth]{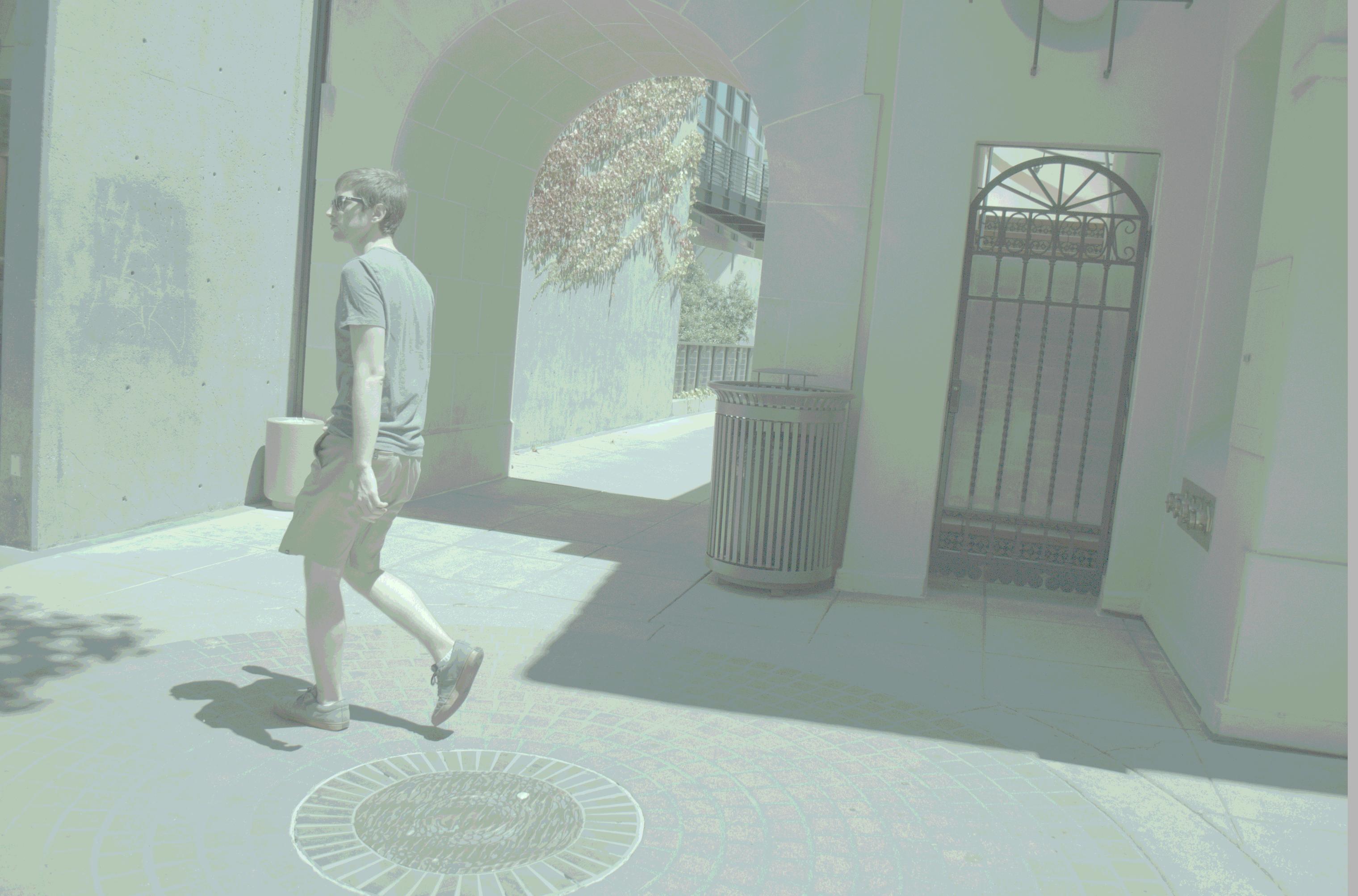}& 
         \includegraphics[width=0.23\linewidth]{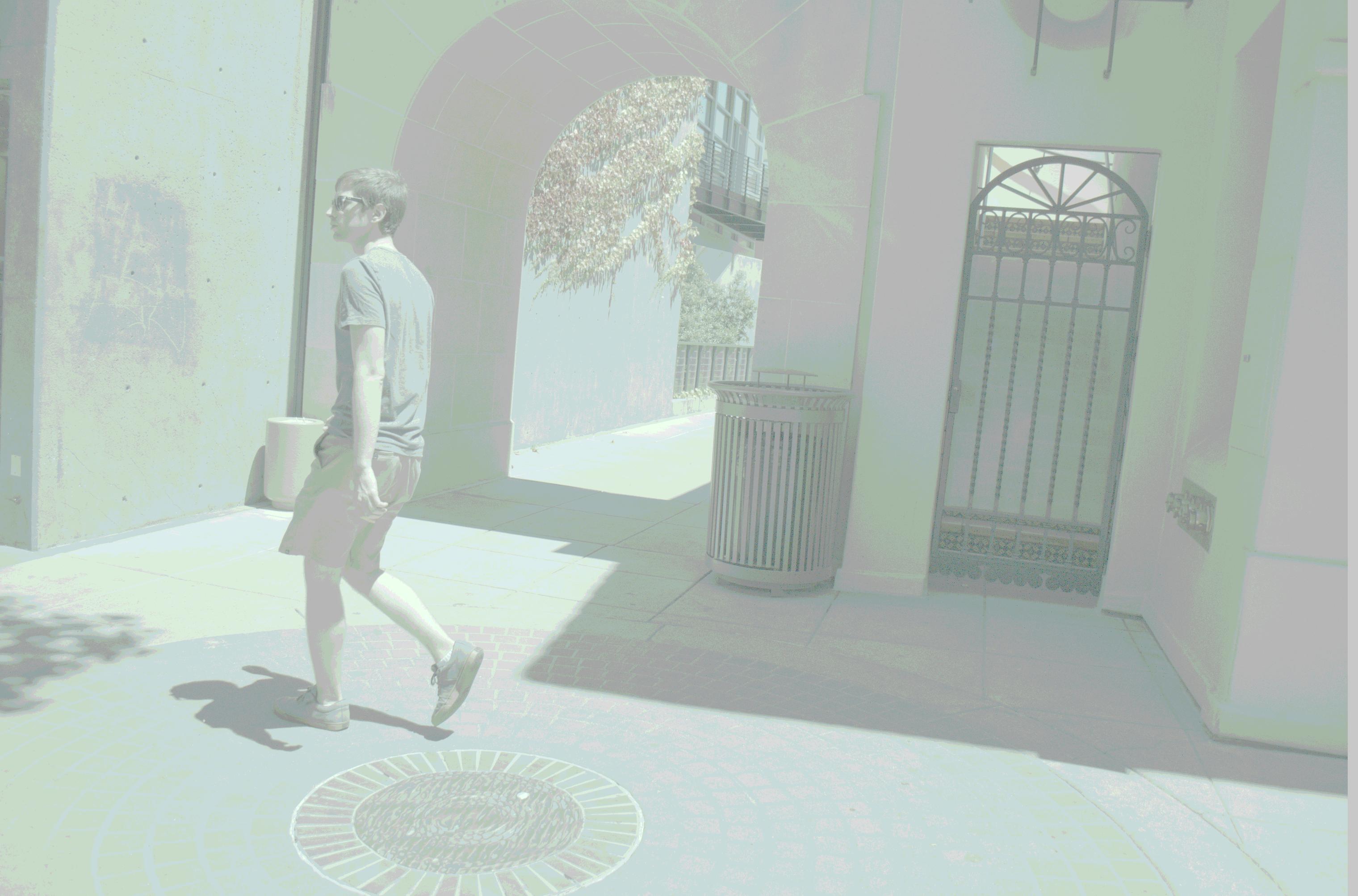}& 
         \includegraphics[width=0.23\linewidth]{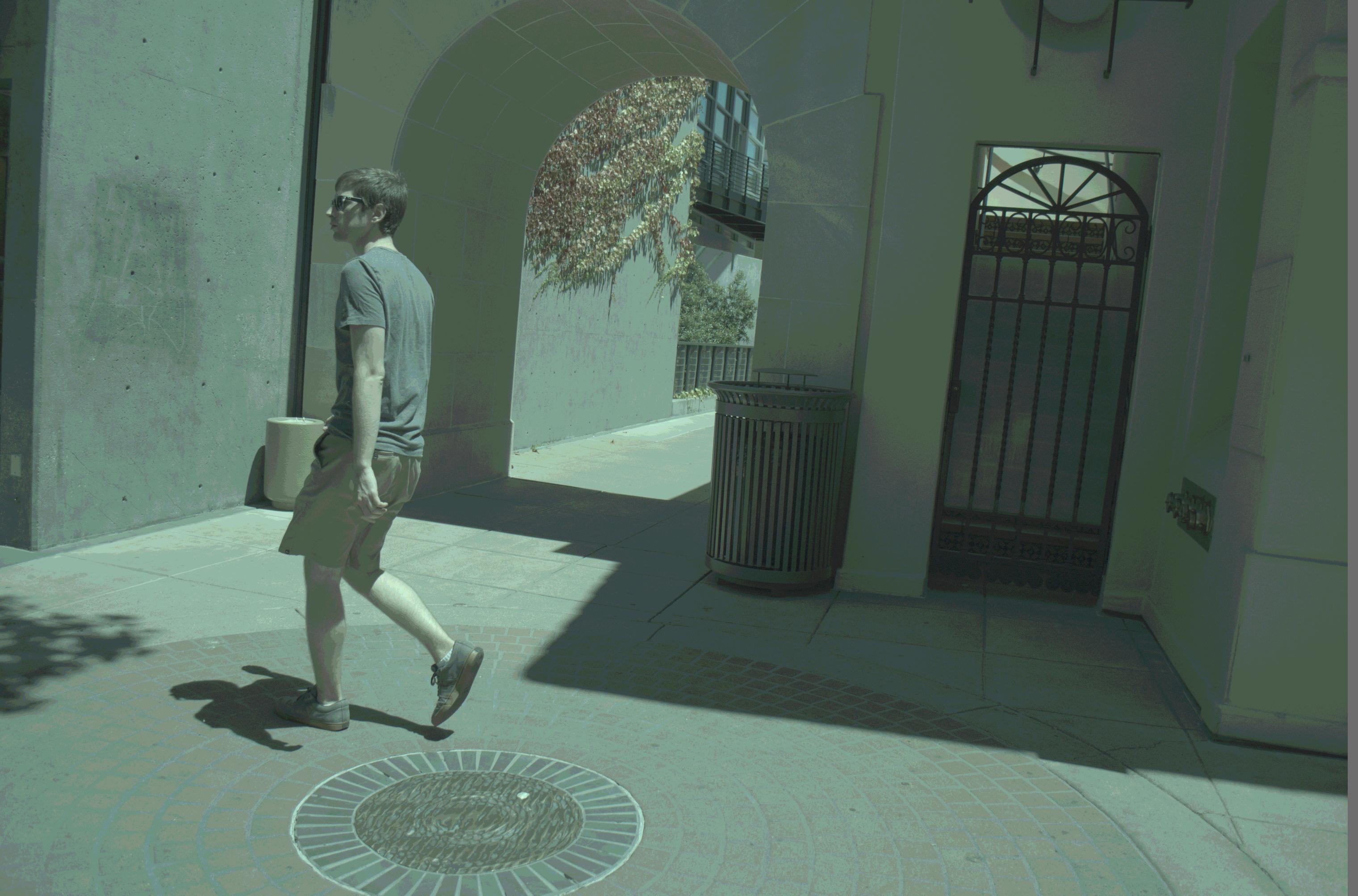}& 
         \includegraphics[width=0.23\linewidth]{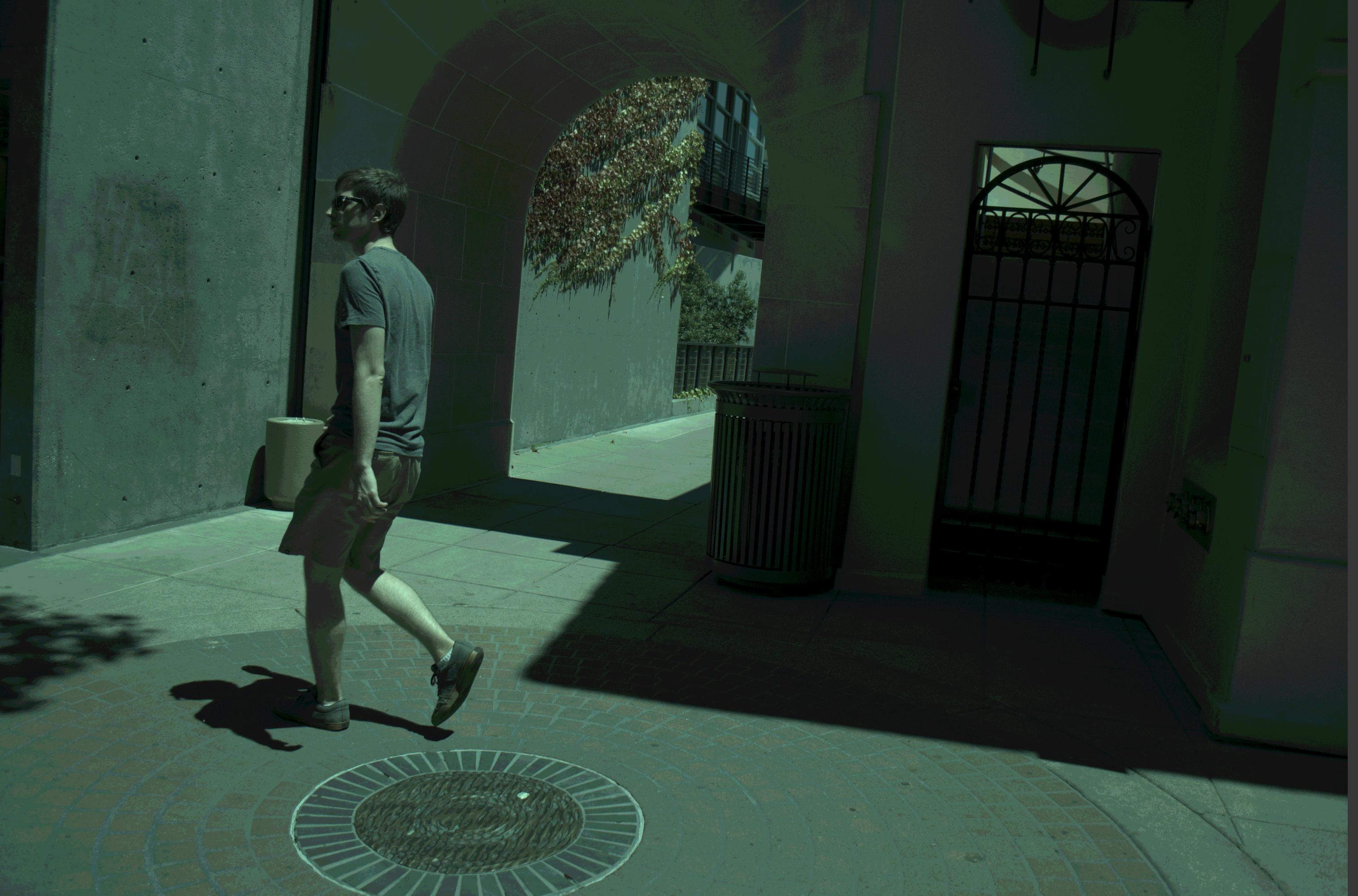}& 
         \includegraphics[width=0.23\linewidth]{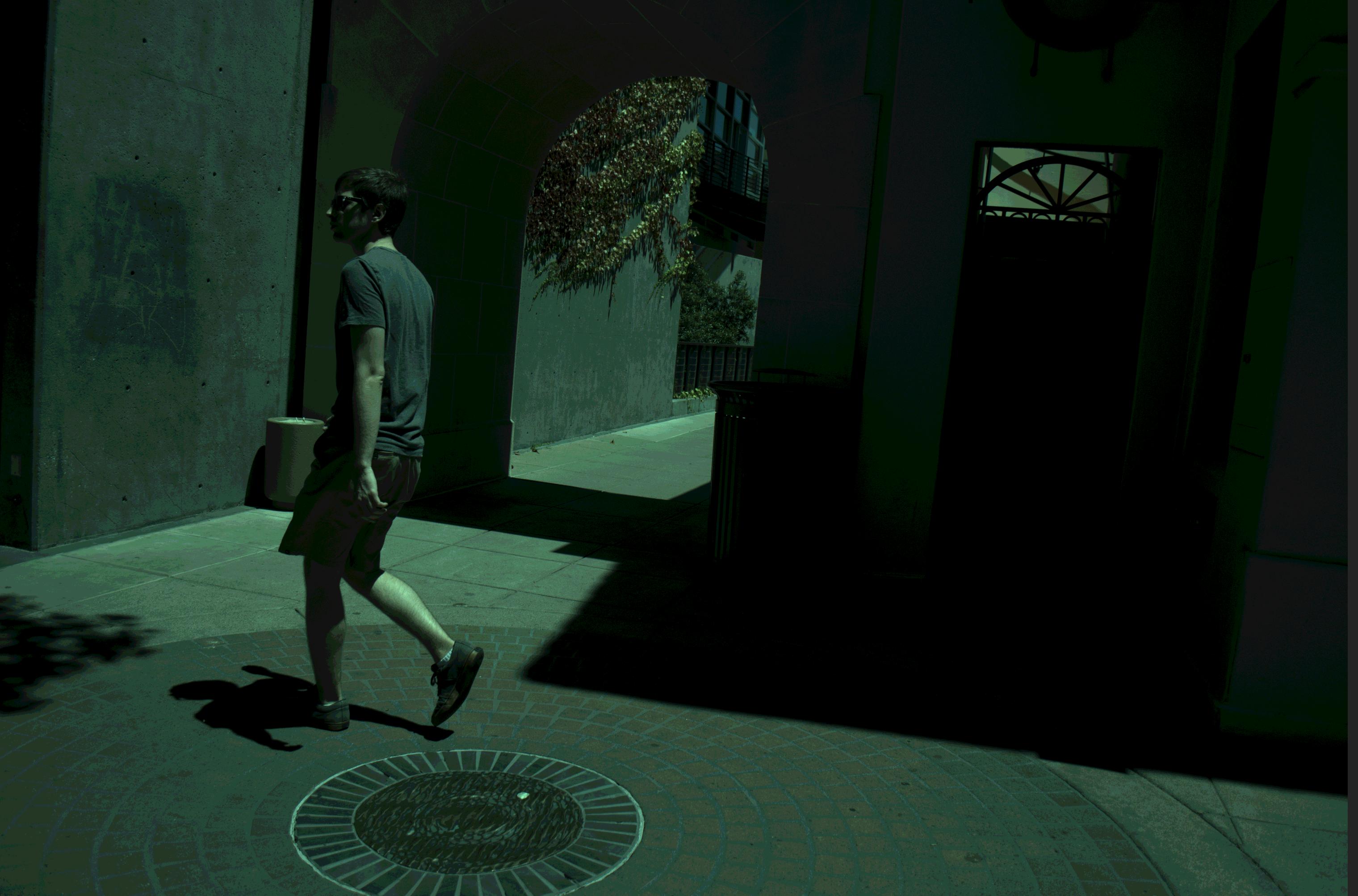}& 
         \includegraphics[width=0.23\linewidth]{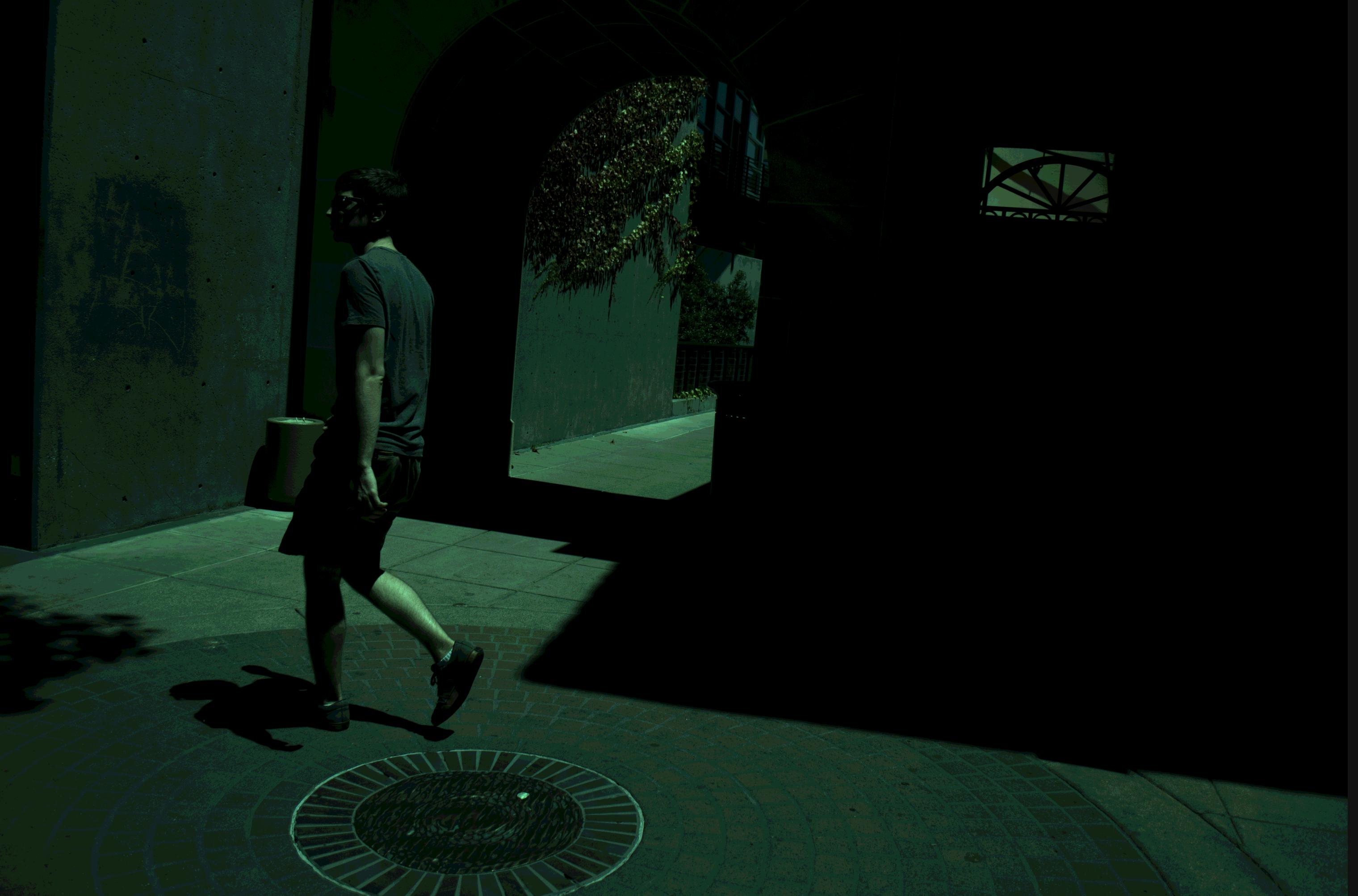}\\

         \rotatebox{90}{\phantom{te}\tiny RAOD Day} & 
         \includegraphics[width=0.23\linewidth]{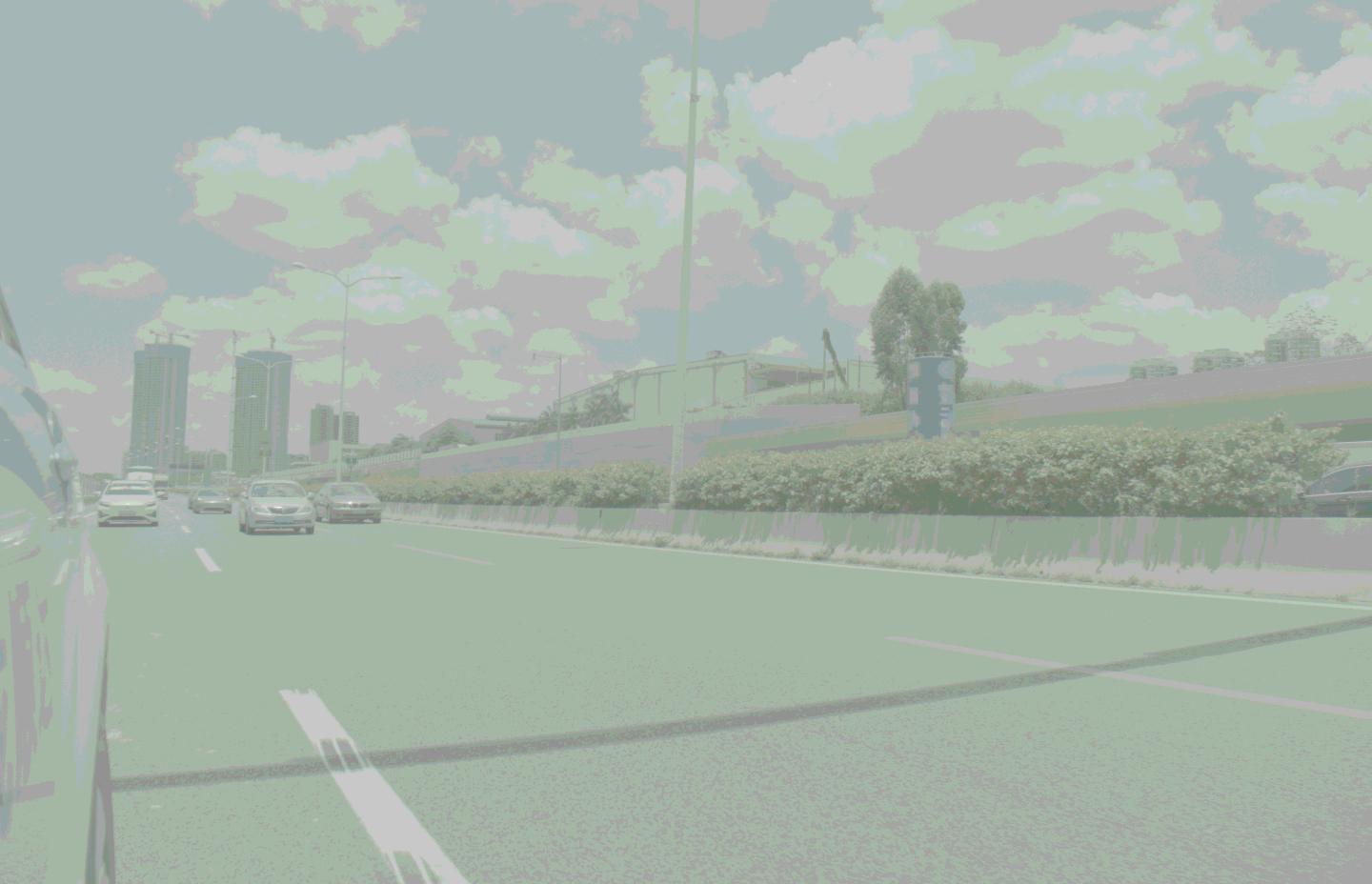}& 
         \includegraphics[width=0.23\linewidth]{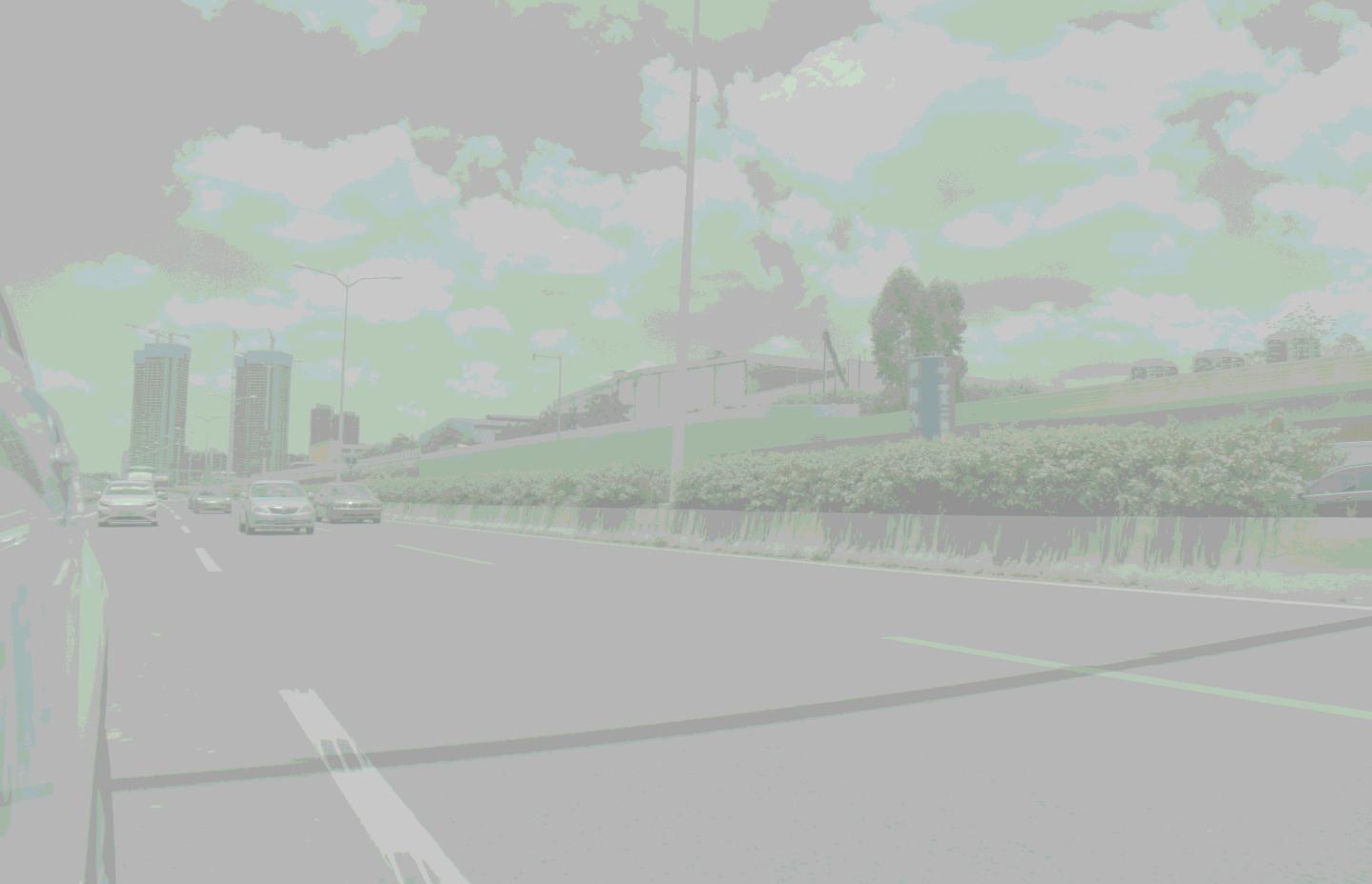}& 
         \includegraphics[width=0.23\linewidth]{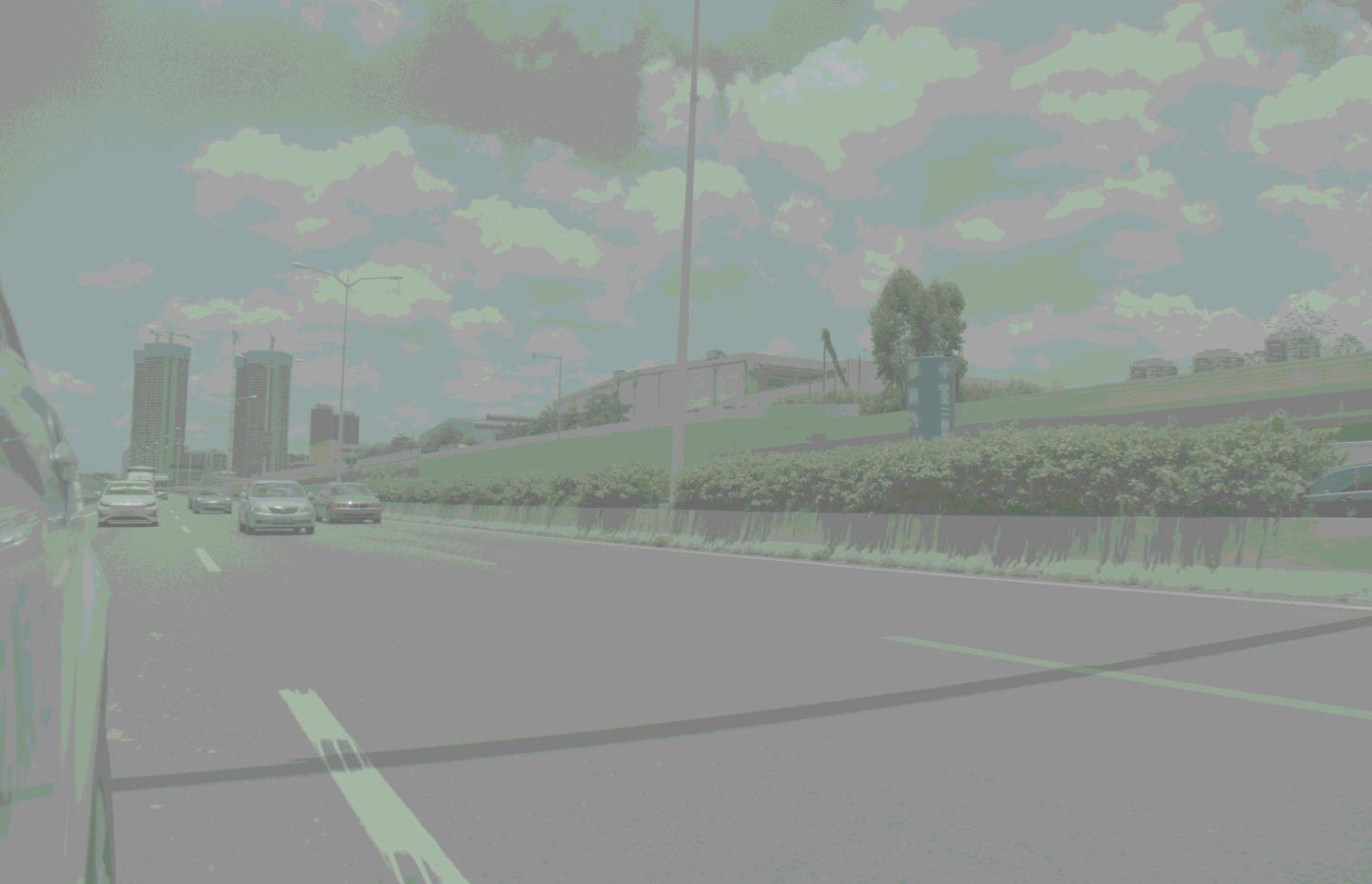}& 
         \includegraphics[width=0.23\linewidth]{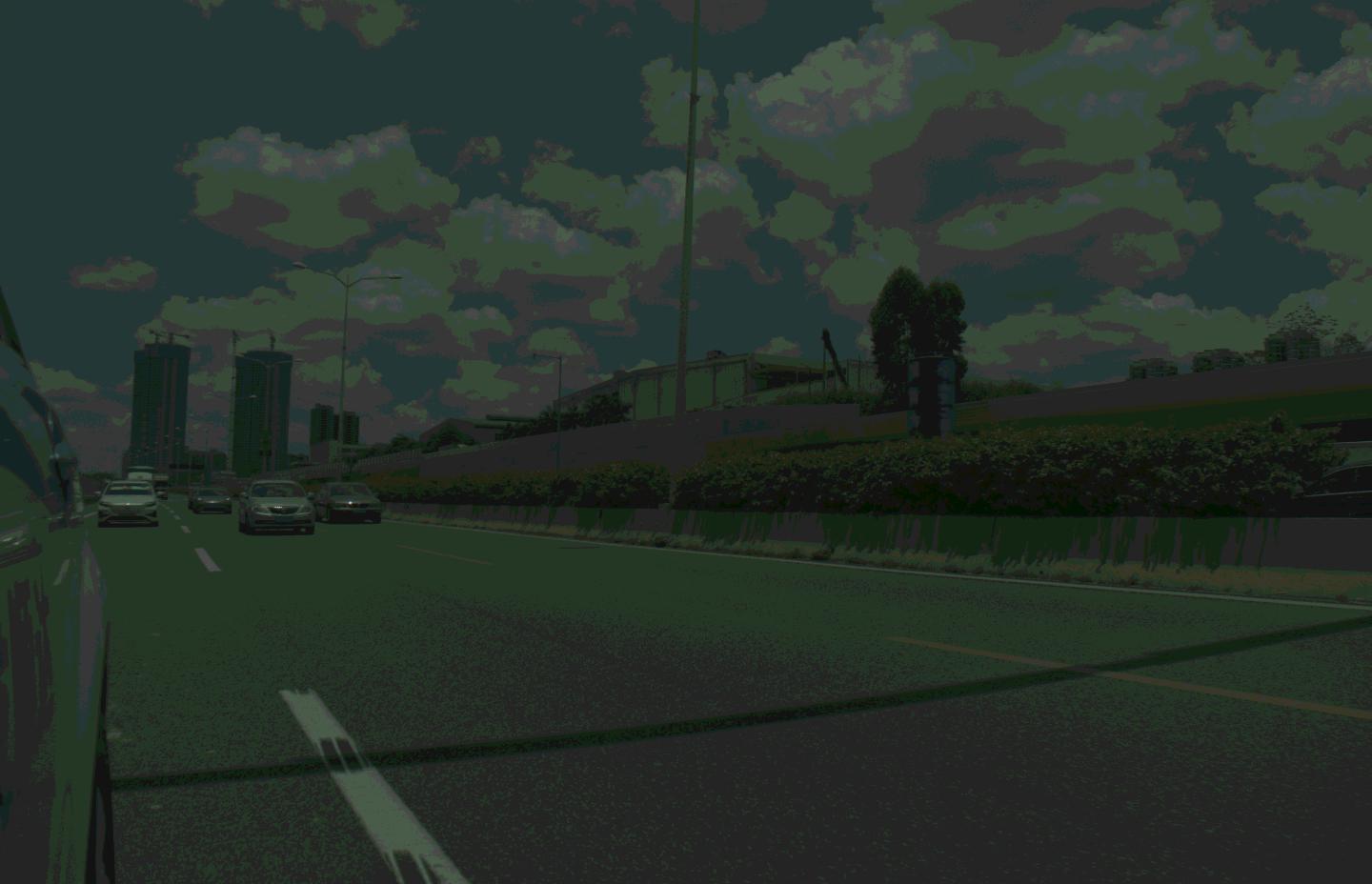}& 
         \includegraphics[width=0.23\linewidth]{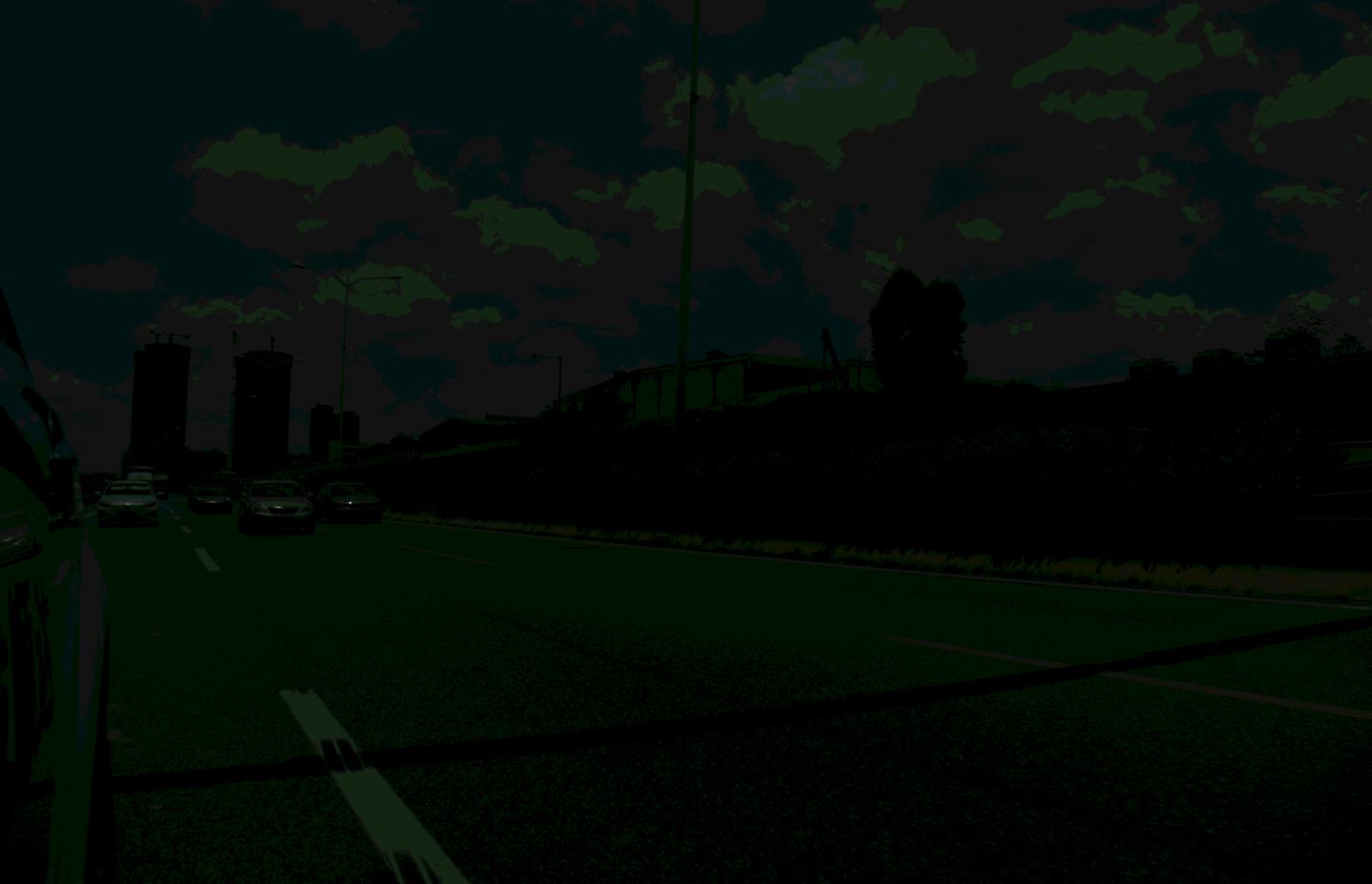}& 
         \includegraphics[width=0.23\linewidth]{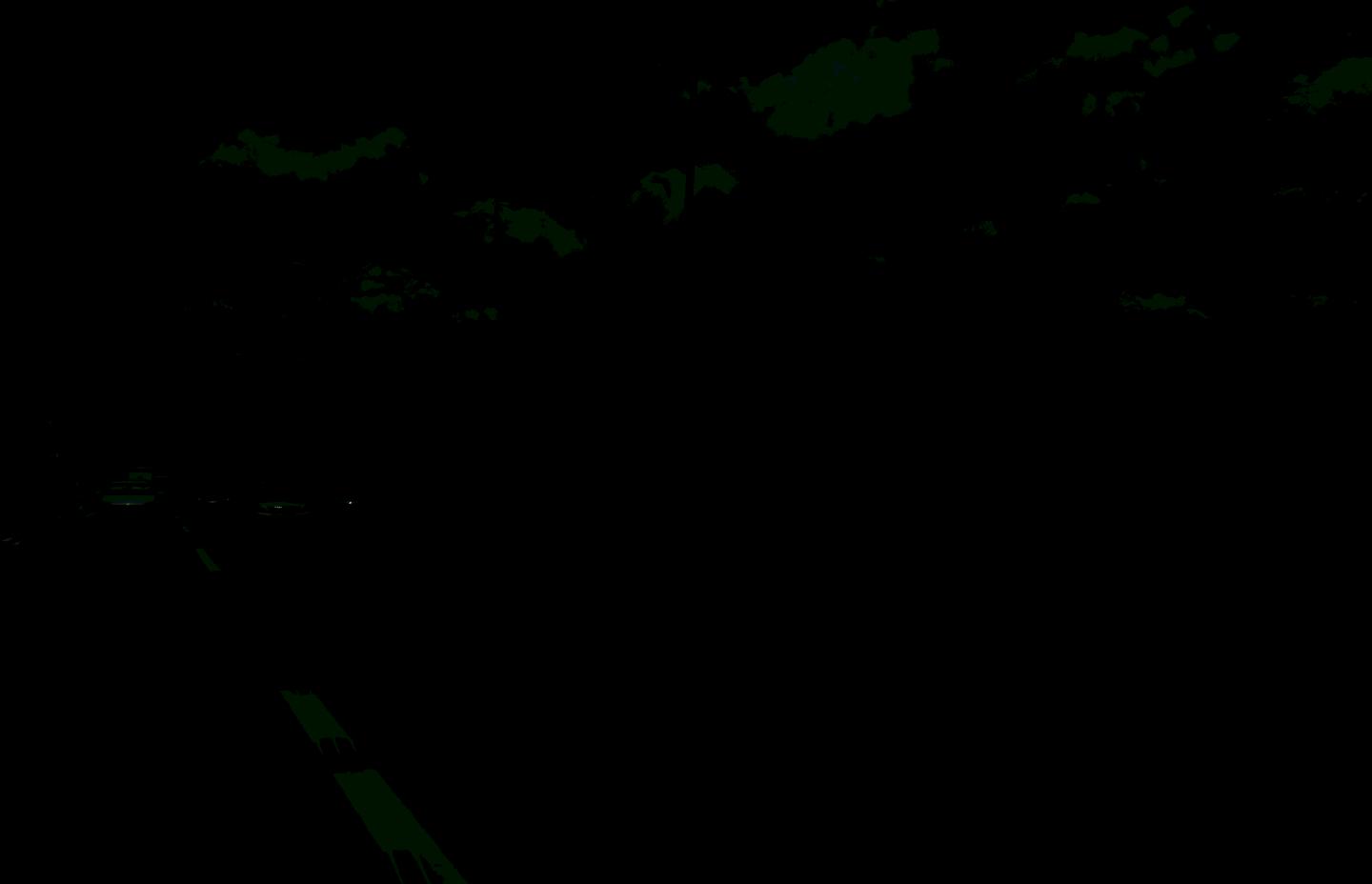}& 
         \includegraphics[width=0.23\linewidth]{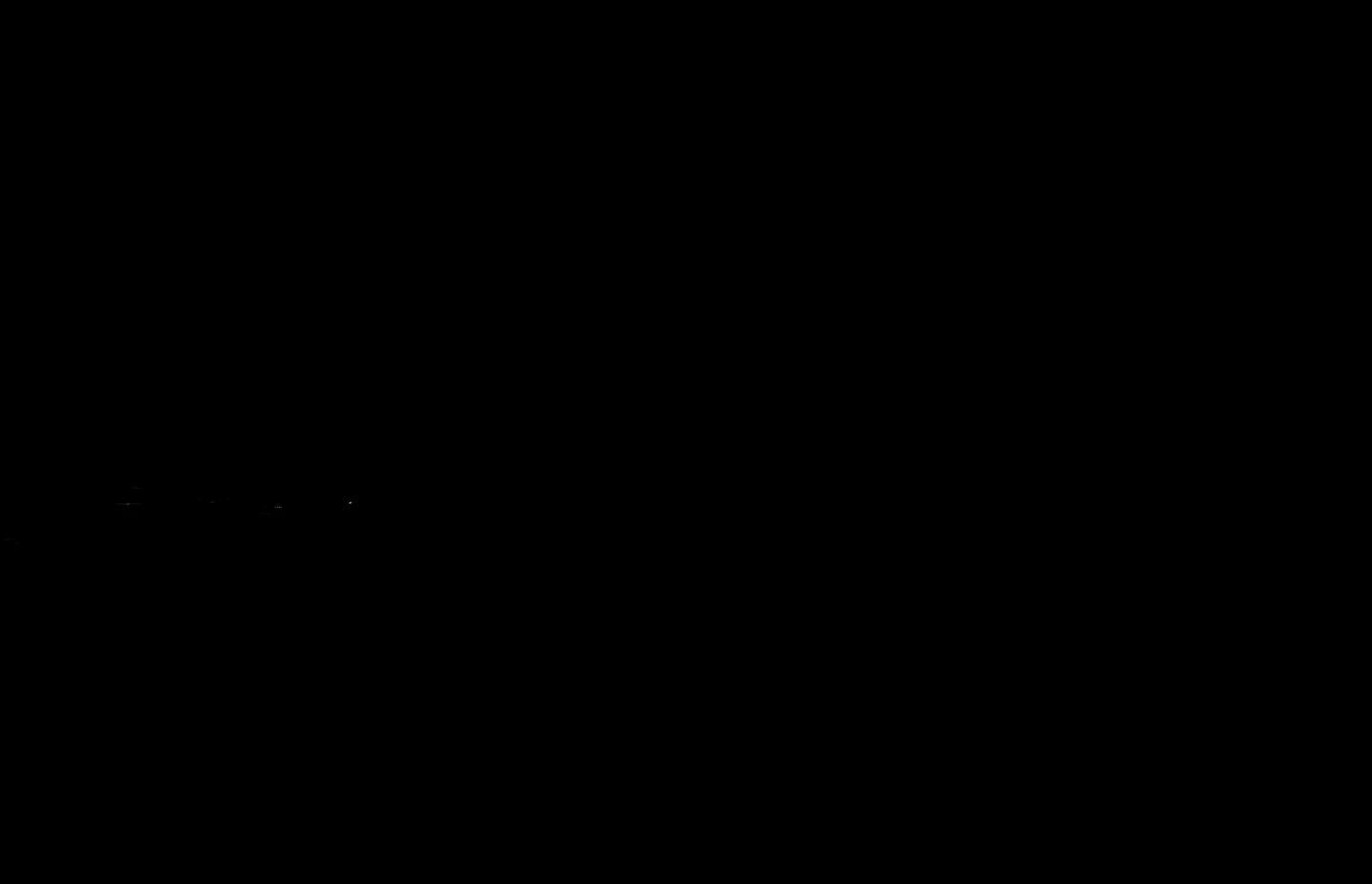}\\

         \rotatebox{90}{\phantom{t}\tiny RAOD Night} & 
         \includegraphics[width=0.23\linewidth]{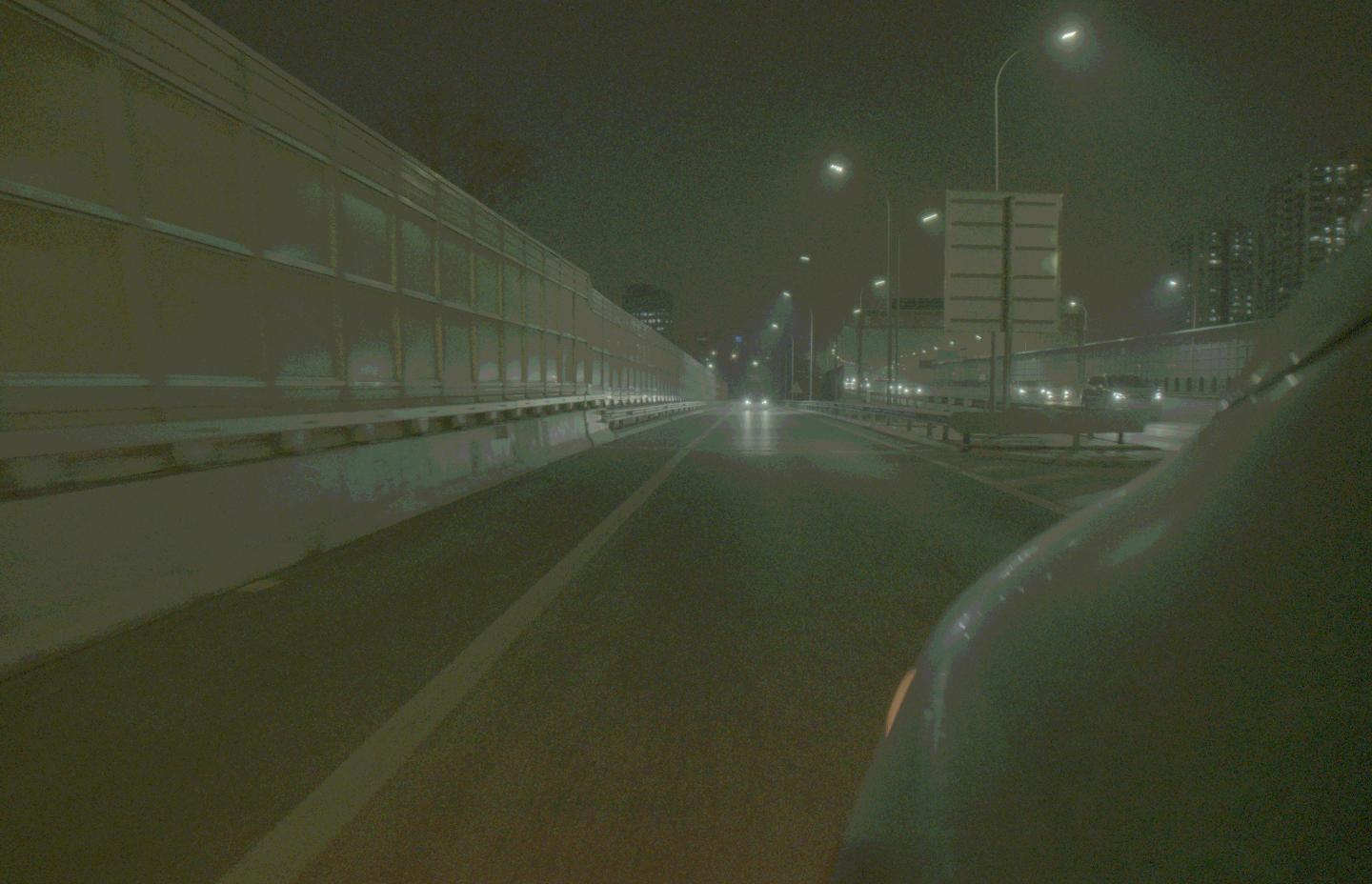}& 
         \includegraphics[width=0.23\linewidth]{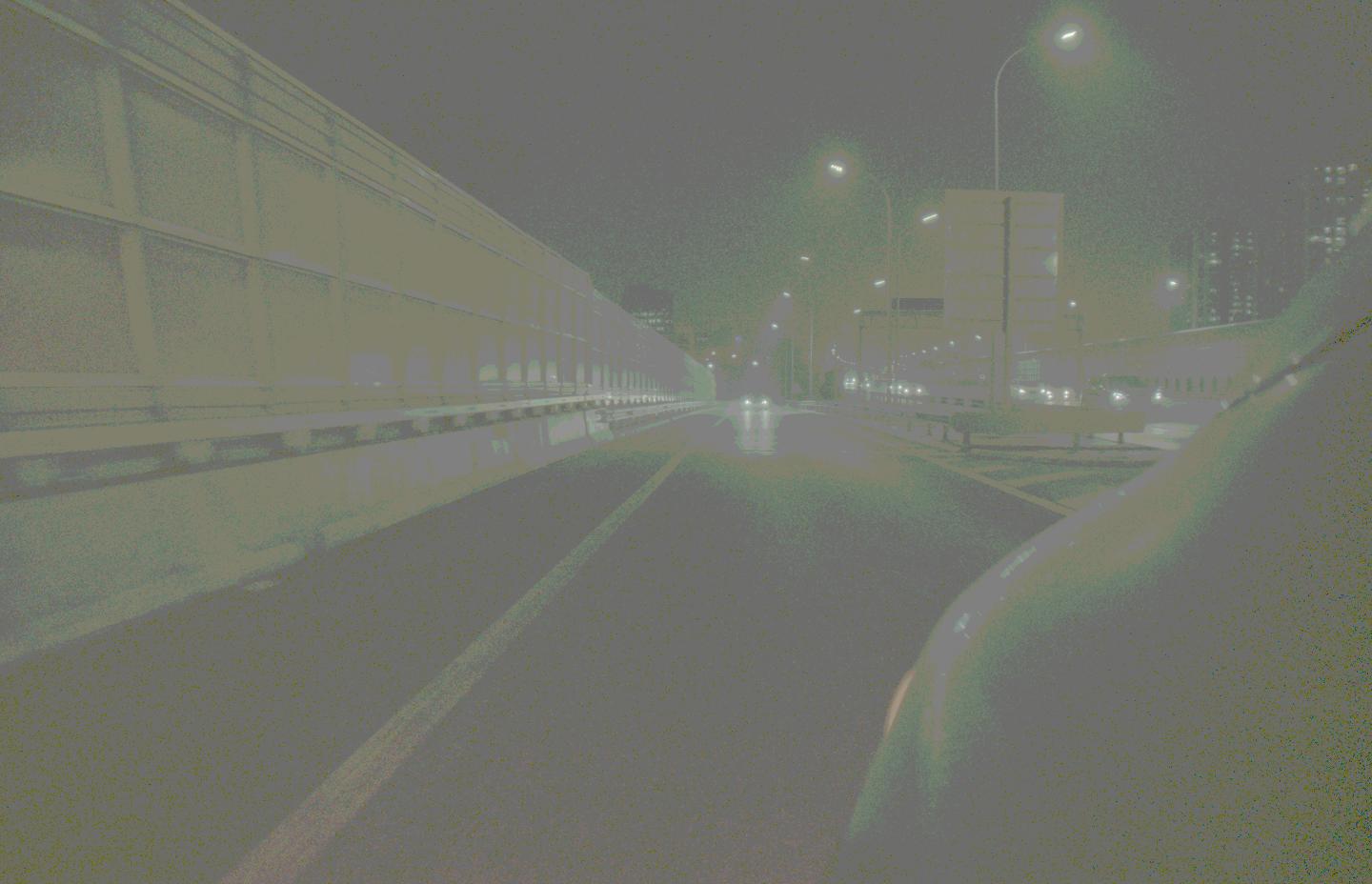}& 
         \includegraphics[width=0.23\linewidth]{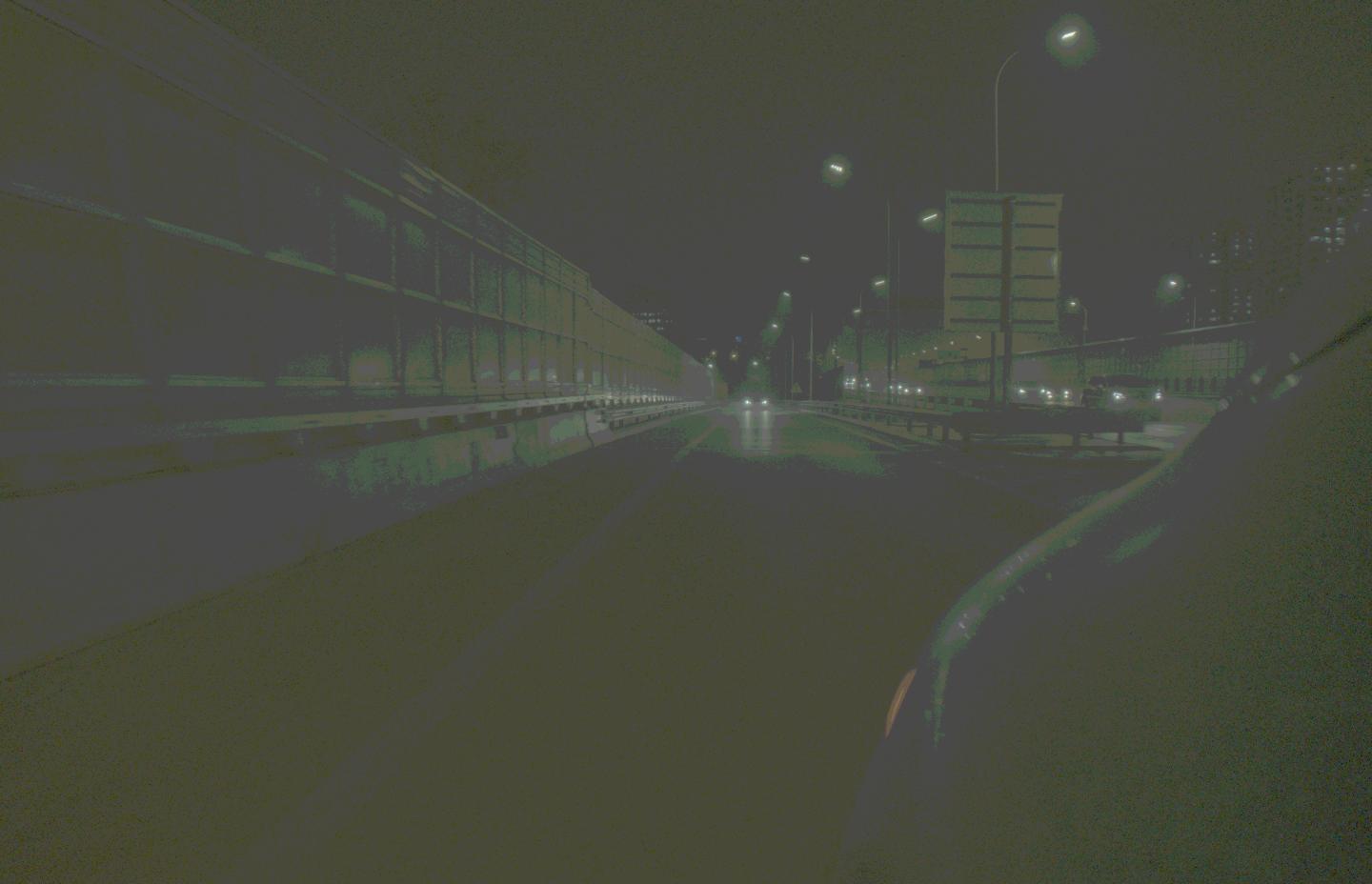}& 
         \includegraphics[width=0.23\linewidth]{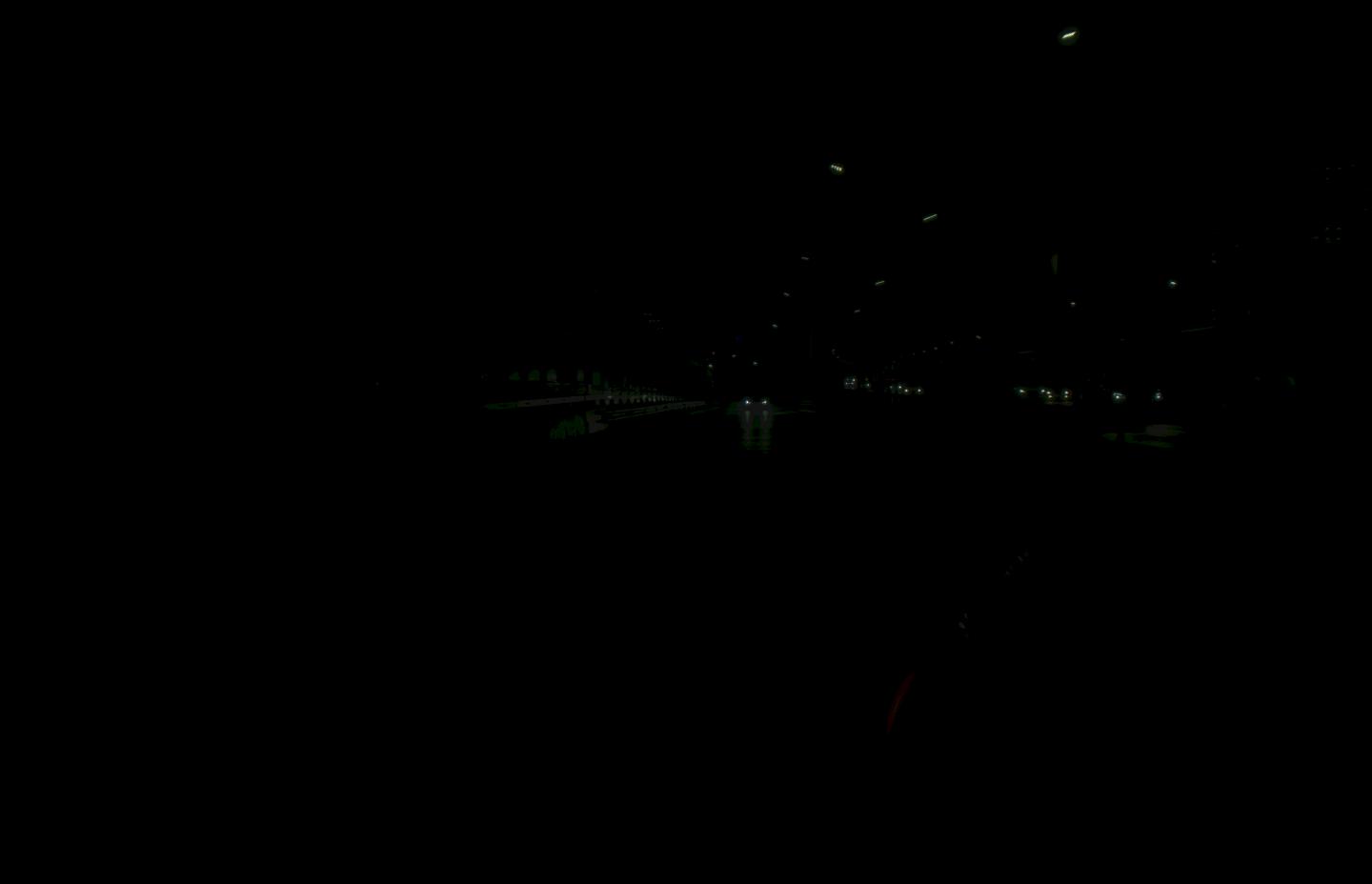}& 
         \includegraphics[width=0.23\linewidth]{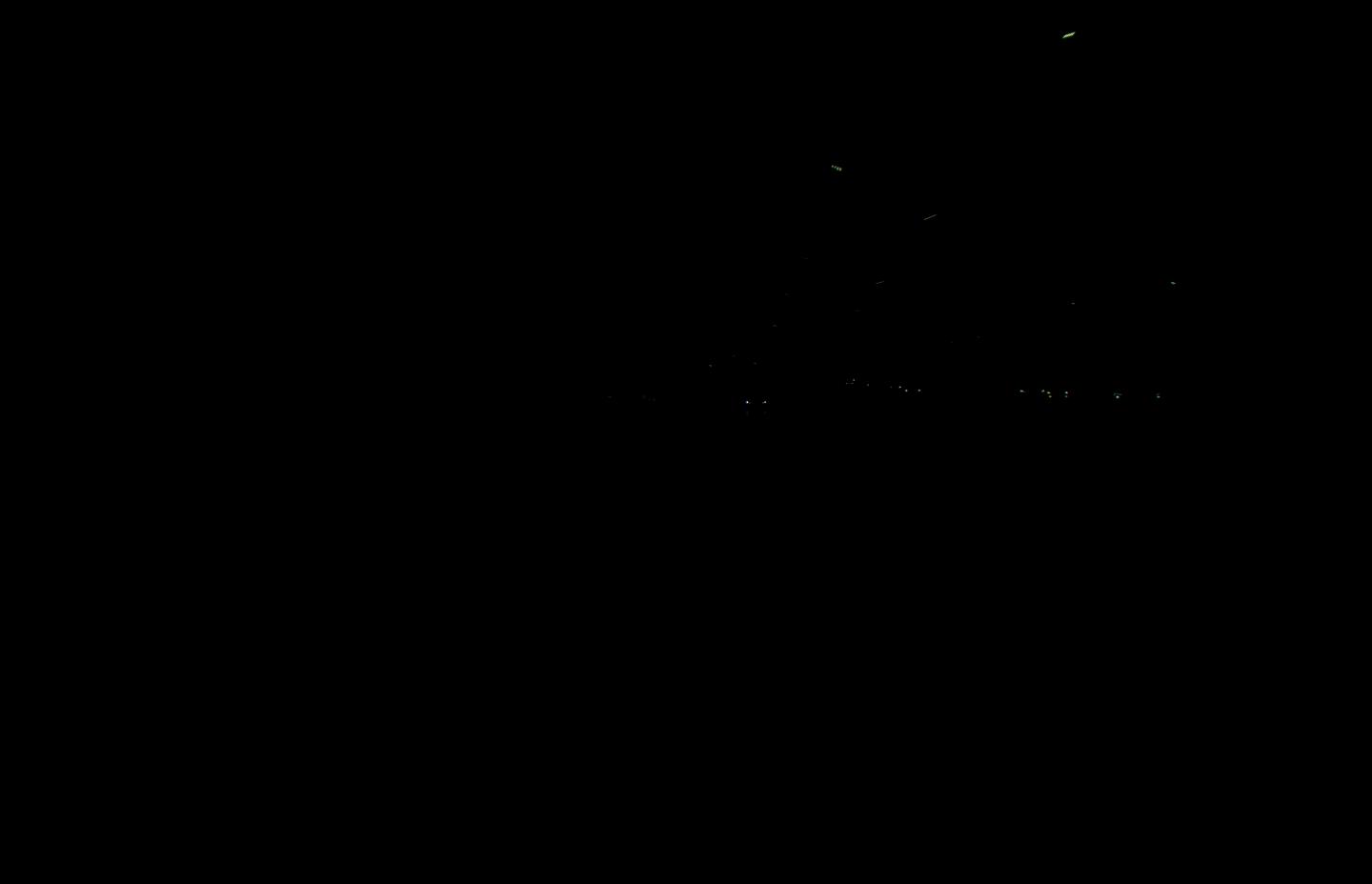}& 
         \includegraphics[width=0.23\linewidth]{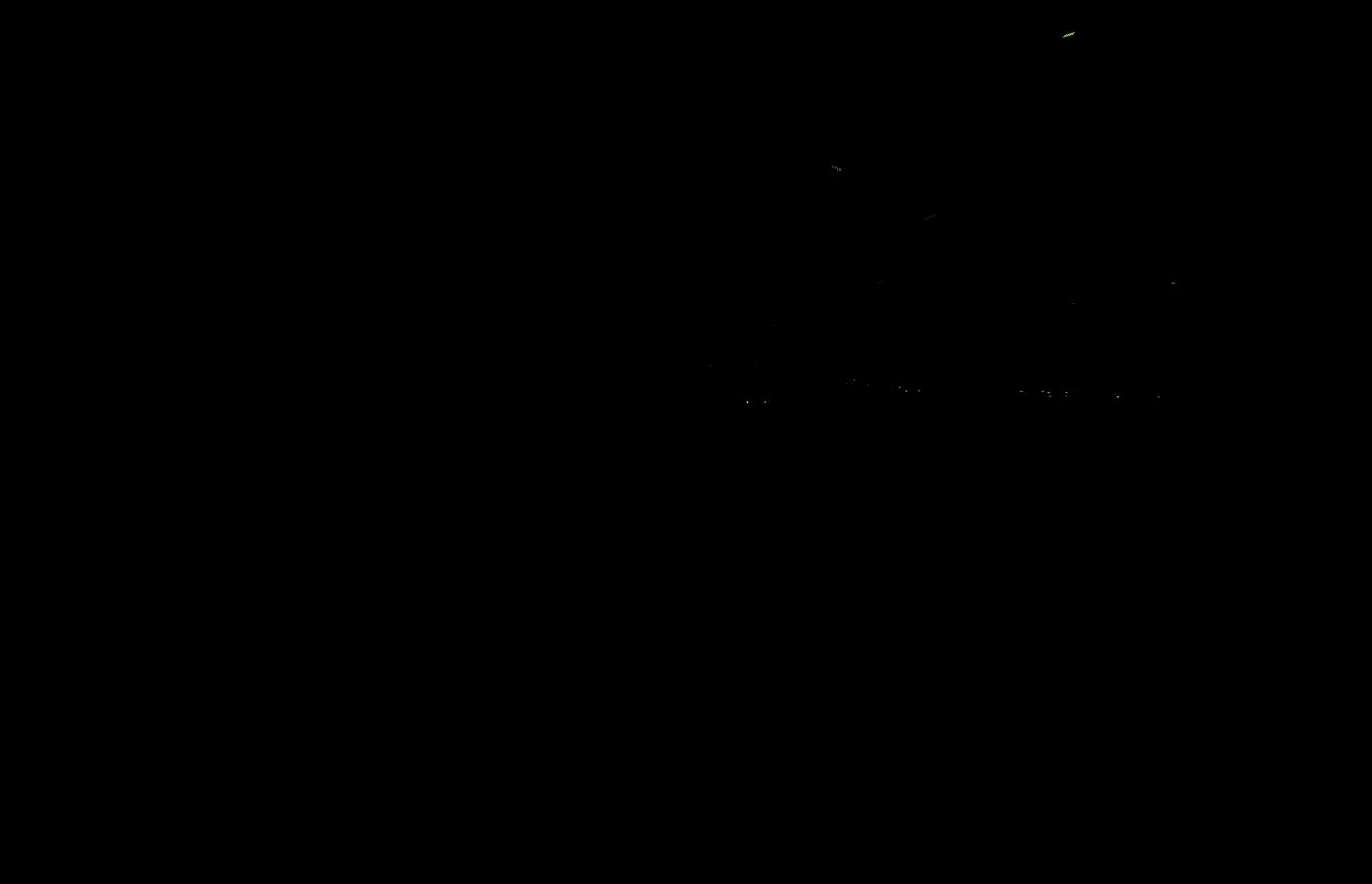}& 
         \includegraphics[width=0.23\linewidth]{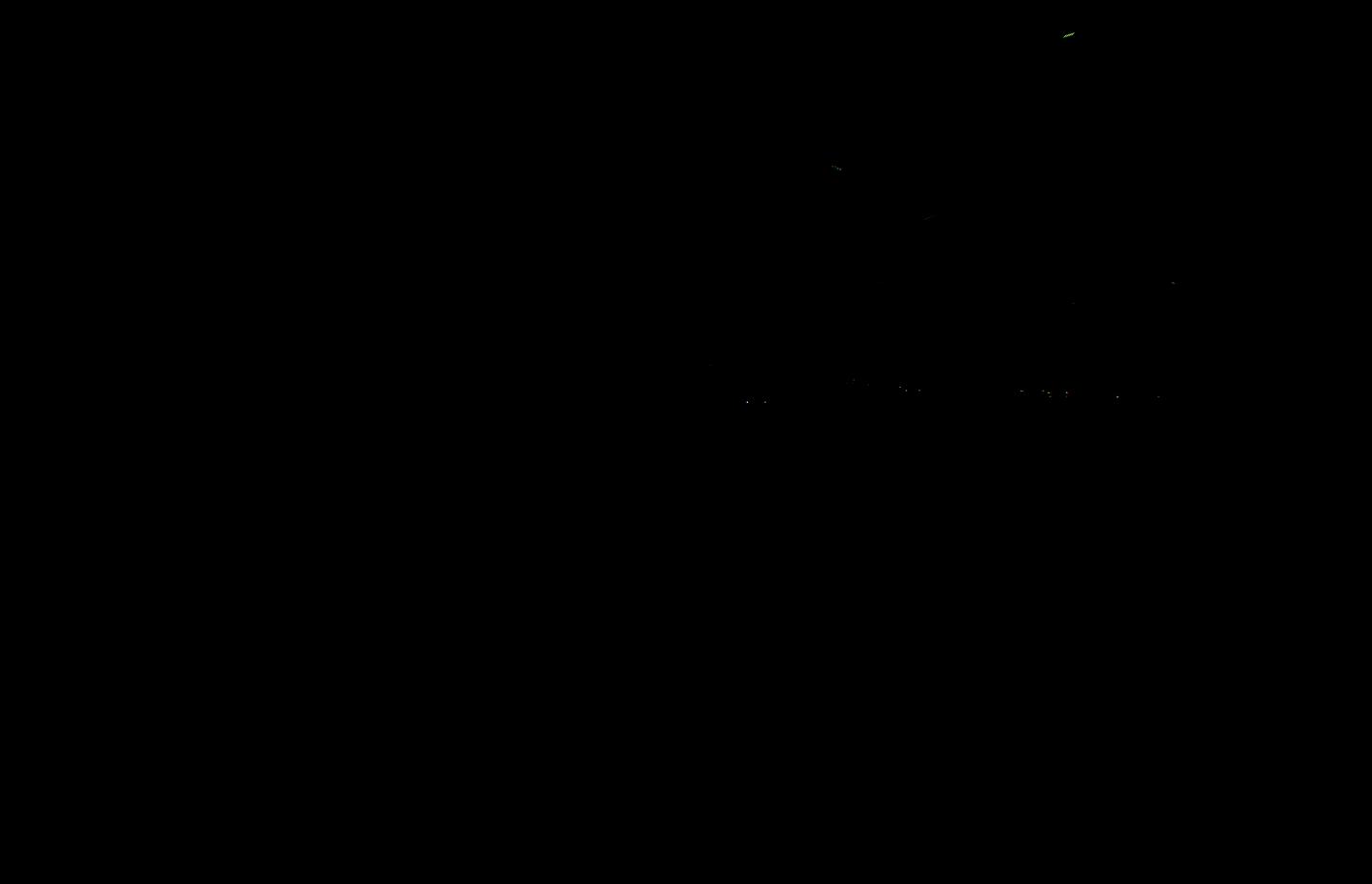}\\
    \end{tabular}
    }
    \caption{We visualize the effect of different values of gamma on a \textbf{4-bit} quantized image. Top row shows images from PASCAL RAW and bottom two rows with RAOD day and night scenarios. The first column applies logarthmic quantization with $\epsilon$=1, whereas the second column approximates log values using appropriate gamma values. The $\gamma$ value for PASCAL RAW is chosen to be 0.136 and 6.395$e$-2 for RAOD. The next five columns show gamma scaling with different chosen gammas. }
    \label{fig:images_processed_different_quant}
\end{figure*}

\section{Additional Results: RAOD Dataset}
\label{sec:appendix:method}
We provide results with RAOD dataset for linear quantization and \method{}. RAOD is a 24 bit HDR RAW data introduced by \cite{xu2023RAODdataset}. \method{} performs best in all cases. Since, RAOD dataset includes day and night scenarios, we also experiment with conditioning gamma on the time of the day. We see that there is marginal difference in the results of the two experiments. 

\begin{figure}[h]
    \centering
    \scalebox{0.9}{
    \begin{tabular}{@{}c@{}}
       \includegraphics[width=1.0\linewidth]{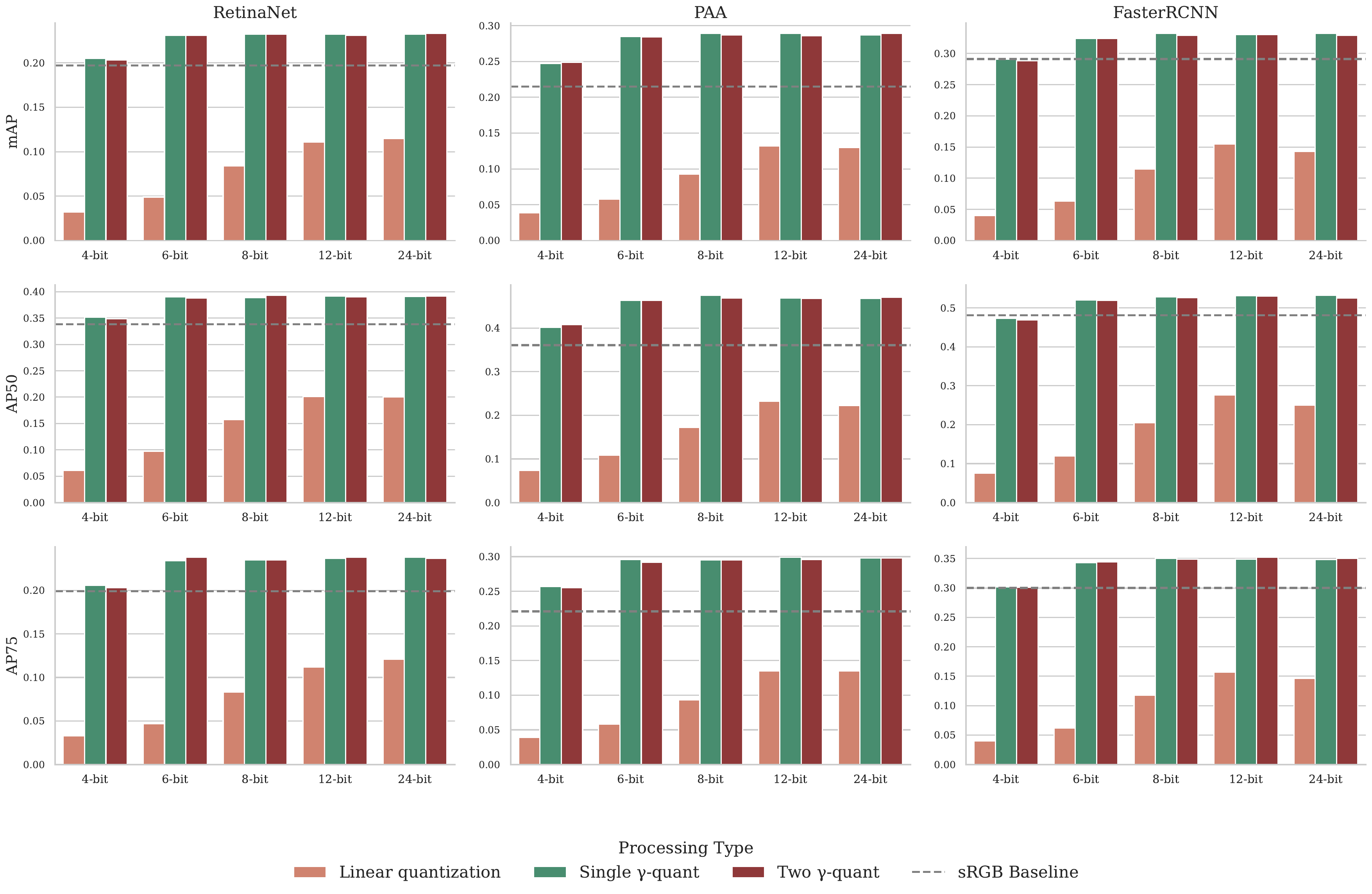}
    \end{tabular}
    }
    \caption{Detection results on the RAOD dataset using \method{} and linear quantization with PAA, Faster R-CNN, and RetinaNet across different bit depths.
    Blue dashed lines represent performance using standard sRGB input images.
    \method{} consistently outperforms standard baselines across all architectures and quantization levels.
    mAP reflects average performance across 10 IoU thresholds; AP50 and AP75 represent results at 0.5 and 0.75 IoU thresholds, respectively.}
    \label{fig:all_mAP_rsults1}
\end{figure}

\section{Additional Results: Gamma Initialization (HAR)}
\label{sec:appendix:gammahar}
In Table~\ref{tab:gammainitinertial} we provide results of comparing a initialization of $\gamma=1$, i.e. linear quantization, with the initialization provided in the paper of $\gamma=0.4$. One can see that results only marginally differ across the two different initializations of $\gamma$, with our chosen initialization of $\gamma=0.4$ slightly performing better, which further supports our assumption made in the main paper that differences in smaller accelerations contain more information than differences in large accelerations.

\begin{table}
\scriptsize
\centering
\caption{Per-dataset HAR results comparing training the chosen initializations of $\gamma=0.4$ with a initialization $\gamma=1.0$, i.e. linear quantization. We provide results for bit depths $\hat{N}=2$ and $\hat{N}=4$. One can see that results only marginally differ with $\gamma=0.4$ performing slightly better, supporting our assumptions made in the main paper.}
\setlength{\tabcolsep}{0.5em}
\label{tab:gammainitinertial}
{\renewcommand{\arraystretch}{1.2}
\begin{tabular}{lc c c c c c}
 &                                     & WEAR            & Wetlab          & Hang-Time       & RWHAR           & SBHAR             \\ \toprule
 \multirow{4}{*}{\rotatebox[origin=c]{90}{$\hat{N} = 4$}}
 & Global ($\gamma=1.0$)               & $70.17 \pm0.13$ & $23.47 \pm0.29$ & $33.89 \pm0.04$ & $70.28 \pm0.10$ & $52.92 \pm0.28$ \\
 & Global ($\gamma=0.4$)               & $70.14 \pm0.51$ & $23.00 \pm0.24$ & $33.63 \pm0.17$ & $69.28 \pm0.46$ & $52.58 \pm0.20$  \\
 & Per-axis ($\gamma=1.0$)             & $68.93 \pm0.41$ & $23.45 \pm0.28$ & $33.82 \pm0.09$ & $70.09 \pm0.91$ & $53.58 \pm0.15$  \\
 & Per-axis ($\gamma=0.4$)             & $70.17 \pm0.39$ & $23.63 \pm0.20$ & $33.67 \pm0.17$ & $71.16 \pm1.26$ & $52.52 \pm0.21$  \\ \midrule
 \multirow{4}{*}{\rotatebox[origin=c]{90}{$\hat{N} = 2$}}
 & Global ($\gamma=1.0$)               & $60.55 \pm0.65$ & $12.98 \pm0.12$ & $29.63 \pm0.12$ & $70.67 \pm0.20$ & $35.75 \pm0.55$  \\
 & Global ($\gamma=0.4$)               & $60.45 \pm0.43$ & $12.94 \pm0.18$ & $30.67 \pm0.12$ & $71.41 \pm0.51$ & $40.18 \pm0.32$  \\
 & Per-axis ($\gamma=1.0$)             & $62.76 \pm0.24$ & $14.35 \pm0.14$ & $28.60 \pm0.26$ & $67.30 \pm0.36$ & $36.56 \pm0.33$  \\
 & Per-axis ($\gamma=0.4$)             & $63.95 \pm0.14$ & $16.85 \pm0.54$ & $30.60 \pm0.22$ & $69.21 \pm0.32$ & $41.88 \pm0.32$ \\ \bottomrule
\end{tabular}
}
\end{table}

\end{document}